\documentclass[
twocolumn,
]{ceurart}

\usepackage{multirow}
\usepackage{tcolorbox}
\usepackage{xcolor,colortbl}
\usepackage{tabularx}
\usepackage{makecell}
\usepackage{siunitx}
\usepackage{booktabs}
\usepackage{float}
\usepackage{graphicx}
\usepackage{caption}
\usepackage{subcaption} 
\usepackage{adjustbox}
\usepackage{enumitem}

\usepackage{listings}
\begin{document}

\copyrightyear{2025}
\copyrightclause{Copyright for this paper by its authors.
  Use permitted under Creative Commons License Attribution 4.0
  International (CC BY 4.0).}

\conference{CLiC-it 2025: Eleventh Italian Conference on Computational Linguistics, September 24 — 26, 2025, Cagliari, Italy}

\title{IMB: An Italian Medical Benchmark for Question Answering}

\author[1,2]{Antonio Romano}
[%
orcid=0009-0000-5377-5051,
email=antonio.romano5@unina.it,
url=https://github.com/LaErre9,
]

\author[1,2]{Giuseppe Riccio}
[%
orcid=0009-0002-8613-1126,
email=giuseppe.riccio3@unina.it,
url=https://github.com/giuseppericcio,
]
\cormark[1]
\author[1,2]{Mariano Barone}
[%
orcid=0009-0004-0744-2386,
email=mariano.barone@unina.it,
url=https://github.com/csmariano,
]

\author[3]{Marco Postiglione}
[%
orcid=0000-0001-6092-940X,
email=marco.postiglione@northwestern.edu,
]

\author[1,2]{Vincenzo Moscato}
[%
orcid=0000-0002-0754-7696,
email=vincenzo.moscato@unina.it,
url=http://wpage.unina.it/vmoscato/,
]

\address[1]{University of Naples Federico II, Department of Electrical Engineering and Information Technology (DIETI), Via Claudio, 21 - 80125 - Naples, Italy}
\address[2]{Consorzio Interuniversitario Nazionale per l'Informatica (CINI) - ITEM National Lab, Complesso Universitario Monte S.Angelo, Naples, Italy}
\address[3]{Northwestern University, Department of Computer Science, McCormick School of Engineering and Applied Science, 2233 Tech Dr, Evanston, IL 60208, United States}

\cortext[1]{Corresponding author.}

\begin{abstract}
Online medical forums have long served as vital platforms where patients seek professional healthcare advice, generating vast amounts of valuable knowledge. However, the informal nature and linguistic complexity of forum interactions pose significant challenges for automated question answering systems, especially when dealing with non-English languages. We present two comprehensive Italian medical benchmarks: \textbf{IMB-QA}, containing 782,644 patient-doctor conversations from 77 medical categories, and \textbf{IMB-MCQA}, comprising 25,862 multiple-choice questions from medical specialty examinations. We demonstrate how Large Language Models (LLMs) can be leveraged to improve the clarity and consistency of medical forum data while retaining their original meaning and conversational style, and compare a variety of LLM architectures on both open and multiple-choice question answering tasks. Our experiments with Retrieval Augmented Generation (RAG) and domain-specific fine-tuning reveal that specialized adaptation strategies can outperform larger, general-purpose models in medical question answering tasks. These findings suggest that effective medical AI systems may benefit more from domain expertise and efficient information retrieval than from increased model scale. We release both datasets and evaluation frameworks in our GitHub repository to support further research on multilingual medical question answering: \url{https://github.com/PRAISELab-PicusLab/IMB}.
\end{abstract}

\begin{keywords}
  Healthcare NLP \sep
  Medical QA Dataset \sep
  Generative AI \sep
  Large Language Models
\end{keywords}

\maketitle

\section{Introduction} \label{sec: introduction}
Since the early days of the Internet, online medical forums have facilitated direct, valuable interactions between patients and healthcare professionals, creating an accessible space for medical advice and support. While these platforms serve as vital resources for medical guidance, they present unique challenges for Natural Language Processing (NLP) systems, particularly in Question Answering (QA) tasks. Unlike traditional medical texts, these conversations are characterized by colloquial language, implicit medical knowledge, and cultural nuances that current QA systems struggle to interpret accurately. Existing biomedical QA research has primarily focused on structured, English-language content, leveraging pre-trained models like BERT \cite{devlin-etal-2019-bert}, RoBERTa \cite{DBLP:journals/corr/abs-1907-11692}, and BioBERT \cite{DBLP:journals/bioinformatics/LeeYKKKSK20}. While these models have shown promising results on standard QA benchmarks \cite{rajpurkar-etal-2016-squad}, \cite{DBLP:conf/nips/YangDYCSL19}, \cite{kwiatkowski-etal-2019-natural}, they are predominantly trained on formal medical literature and standardized exam questions \cite{DBLP:journals/bmcbi/TsatsaronisBMPZ15}. This creates a significant gap between model capabilities and real-world medical communication needs, particularly in non-English contexts. To address these challenges, we introduce two complementary datasets: \textbf{IMB-QA} (Italian Medical Benchmark for Question Answering), a comprehensive collection of 782,644 real-world medical conversations across 77 medical categories from Italian online forums MedicItalia\footnote{\url{https://www.medicitalia.it/}} and Dica33\footnote{\url{https://www.dica33.it/}}; and \textbf{IMB-MCQA} (Italian Medical Benchmark for Multiple Choice Question Answering), containing 25,862 multiple-choice questions and answers from medical specialty admission exams collected from the simulator CompitoInClasse.org\footnote{\url{https://www.compitoinclasse.org/}}. Both datasets have been carefully curated, with \textbf{IMB-QA} specifically enhanced through LLM-based methodologies to ensure quality and anonymity while preserving the authentic nature of patient-doctor interactions.

Our work goes beyond data contribution through extensive experimentation with state-of-the-art language models. We conduct a systematic evaluation of various LLM architectures, comparing models of different sizes and training backgrounds, with particular attention to those specialized in biomedical domains. Through this analysis, we explore the two standard approaches to enhance medical QA performance: Retrieval Augmented Generation (RAG) and in-domain fine-tuning. Our experiments with RAG demonstrate significant improvements in response accuracy and completeness, while our fine-tuning studies reveal the potential of task adaptation even for smaller models. The dual nature of our datasets --- spanning both informal forum discussions and formal medical examinations --- provides a unique opportunity to assess model performance across different types of medical communication. Our findings challenge conventional assumptions about model size and generalization, suggesting that targeted task adaptation and retrieval-based approaches may be more crucial for medical QA than raw model scale.

\section{Related work} \label{sec: related_work}
In Question Answering (QA), models are typically provided with a relevant text from which they must extract answers. However, in real-world applications, manually curating such texts is impractical due to the high cost of obtaining annotated contexts. This challenge has driven the development of Open-Domain QA (OpenQA), where models must autonomously retrieve and understand relevant information to generate accurate responses \cite{DBLP:journals/tacl/WangHJ024}. In the biomedical domain, numerous datasets have been introduced to advance QA, particularly in high-resource languages such as English (as shown in Table \ref{tab:qa_mcqa_comparison}). However, resources for other linguistic domains—especially Italian—remain scarce, limiting the development and evaluation of multilingual biomedical QA models.
\begin{table}[t]
    \centering
    \caption{Comparison of QA and MCQA datasets from prior literature and our proposed \textbf{IMB} datasets.}
    \scalebox{0.90}{
    \begin{tabular}{llrl}
        \toprule
        \textbf{Type} & \textbf{Dataset} & \textbf{\# Q/A} & \textbf{Language} \\
        \midrule
        \multirow{12}{*}{\textbf{QA}} 
        & BiQA \cite{BiQA} & $>$7.4K & English \\
        & HealthQA \cite{HealthQA} & $>$7.5K & English \\
        & EPIC-QA \cite{Epic-QA} & 45 & English \\
        & COVID-QA \cite{moller2020covidqa} & $>$2K & English \\
        & CliCR \cite{suster-daelemans-2018-clicr} & $>$100K & English \\
        & LiveQA-Med \cite{LiveMedQA2017} & 738 & Multilingual \\
        & PubMedQA \cite{medqa} & $>$212K & English \\
        & emrQA \cite{pampari2018emrqa} & $>$455K & English \\
        & webMedQA \cite{webmedqa} & $>$63K & English \\
        & BioASQ \cite{bioasq} & $>$3.2K & English \\
        & \cellcolor{cyan!25} \textbf{IMB-QA (Ours)} & \cellcolor{cyan!25} $>$782K & \cellcolor{cyan!25} Italian \\
        \midrule
        \multirow{7}{*}{\textbf{MCQA}} 
        & HEAD-QA \cite{vilares-gomez-rodriguez-2019-head} & $>$6.8K & Spanish \\
        & MedMCQA \cite{MedMCQA} & $>$194K & English \\
        & cMedQA \cite{medqa} & $>$54K & Chinese \\
        & ChiMed \cite{tian-etal-2019-chimed} & $>$24.9K & Chinese \\
        & MEDQA \cite{medqa} & $>$61K & English-Chinese \\
        & QA4-MRE \cite{Peas2013QA4MRE2O} & >1.5K & Multilingual \\
        & \cellcolor{cyan!25} \textbf{IMB-MCQA (Ours)} & \cellcolor{cyan!25} $>$25K & \cellcolor{cyan!25} Italian \\
        \bottomrule
    \end{tabular}}
    \label{tab:qa_mcqa_comparison}
\end{table}

\paragraph{Open-Domain and MRC Biomedical QA}
Several datasets support OpenQA and Machine Reading Comprehension (MRC) in the biomedical field. BiQA \cite{BiQA} compiles questions from online forums (e.g., Stack Exchange, Reddit) and links them to PubMed articles, though the accuracy of this linking remains largely unverified. HealthQA \cite{HealthQA} consists of manually curated medical questions with answers sourced from patient information websites, yet it lacks a systematic quality assessment. BioRead \cite{pappas-etal-2018-bioread} and its extended version, BioMRC \cite{pappas-etal-2020-biomrc}, annotate texts using Unified Medical Language System (UMLS) concepts, enhancing knowledge representation but focusing more on structured information extraction rather than OpenQA. The COVID-19 pandemic and the creation of specialized datasets such as EPIC-QA \cite{Epic-QA} and COVID-QA \cite{moller2020covidqa}, which compile question-answer pairs from pandemic-related literature. However, their long-term relevance is inherently limited to this specific context. CliCR \cite{suster-daelemans-2018-clicr} employs cloze-style questions derived from clinical case reports to assess comprehension and inference abilities, yet its scope is restricted to a narrow set of medical conditions. Although most biomedical QA datasets are available only in English, some efforts have targeted other languages. LiveQA-Med \cite{LiveMedQA2017} provides a small set of 634 annotated medical question-answer pairs, but its test set (104 questions) is too limited for robust evaluation. MEDQA \cite{medqa}, built from medical board exams in English and Chinese, does not clearly specify the balance between languages or the translation quality. WebMedQA \cite{webmedqa}, derived from Chinese health consultancy platforms, reflects real-world medical inquiries, though its reliability depends on the moderation of user-generated content.

\paragraph{Multiple Choice QA}
Several datasets focus on multiple-choice QA (MCQA) for biomedical applications. HEAD-QA \cite{vilares-gomez-rodriguez-2019-head} and MedMCQA \cite{MedMCQA} assess domain knowledge and reasoning skills but lack coverage for Italian. PubMedQA presents a distinct format where article titles serve as binary-answer questions, though it does not address complex inferential reasoning. While ChiMed \cite{tian-etal-2019-chimed} and cMedQA \cite{medqa} provide Chinese-language biomedical MCQA datasets, Italian biomedical QA resources remain virtually nonexistent. QA4-MRE \cite{Peas2013QA4MRE2O} attempted to introduce multilingual medical reading comprehension, yet its dataset was limited in both scale and scope. To address this gap, we introduce a large-scale Italian biomedical QA dataset, consisting of 782,644 question-answer pairs spanning 77 medical categories, alongside an Italian biomedical MCQA dataset with 25,862 multiple-choice questions across 60 categories. Compared to existing datasets, our corpus is significantly larger and more diverse, enhancing both domain-specific knowledge extraction and OpenQA capabilities. Furthermore, we employ advanced post-processing techniques to improve answer accuracy and applicability in medical information retrieval tasks.

\section{IMB Dataset} \label{sec: data}
The IMB dataset consists of two structured subsets: \textbf{IMB-QA}, which focuses on unstructured, patient-driven medical inquiries and professional responses, and \textbf{IMB-MCQA}, which contains structured multiple-choice questions designed for evaluating domain-specific medical knowledge. The \textbf{IMB-QA} dataset captures natural, patient-driven inquiries and professional responses, reflecting real-world medical concerns and interactions (refer to Table \ref{tab:psychotherapy_specialization} for an example).

\begin{table}[t]
    \centering
    \caption{Example of a question-answer pair from \textbf{IMB-QA}.}
    \scalebox{0.76}{
    \begin{tabular}{p{2.5cm} p{6cm}}
        \hline
        \multicolumn{2}{c}{\textbf{IMB-QA}} \\
        \hline
        \textbf{ID} & 14571 \\
        \hline
        \textbf{URL} & \href{https://www.medicitalia.it/consulti/psicoterapia/23549-psichiatri-per-fare-psicoterapia-necessitano-di-specializzazione.html}{https://www.medicitalia.it/consulti/psicoterapia/23549-psichiatri-per-fare-psicoterapia-necessitano-di-specializzazione.html} \\
        \hline
        \textbf{Question} &  
        \textbf{ITA:} Psichiatri: per fare psicoterapia necessitano di specializzazione?  
        Gentili dottori, uno psichiatra è automaticamente abilitato a fare lo psicoterapeuta, o deve, come gli psicologi che intendono anche essere psicoterapeuti, fare i 4 anni di specializzazione in psicoterapia? Grazie.  
        
        \textbf{ENG:} Psychiatrists: do they need specialization to practice psychotherapy?  
        Dear doctors, is a psychiatrist automatically allowed to practice psychotherapy, or does he need, like psychologists who also wish to be psychotherapists, to undergo 4 years of specialization in psychotherapy? Thank you. \\
        \hline
        \textbf{Reformulated Answer} &  
        \textbf{ITA:} Dopo la specializzazione, uno psichiatra può iscriversi all'albo degli psicoterapeuti senza dover completare un'ulteriore specializzazione.  
        Sebbene alcuni psichiatri scelgano di proseguire con studi supplementari in psicoterapia, tale formazione non è obbligatoria.  
        Gli psicologi, invece, devono frequentare un corso di specializzazione per diventare psicoterapeuti.  
        
        \textbf{ENG:} After specialization, a psychiatrist can register with the list of psychotherapists without needing further specialization.  
        While some psychiatrists choose to pursue additional studies in psychotherapy, this training is not mandatory.  
        Psychologists, however, must attend a specialization course to become psychotherapists. \\
        \hline
        \textbf{Category} &  
        \textbf{ITA:} Psicoterapia - \textbf{ENG:} Psychotherapy \\
        \hline
        \textbf{Macro Category} &  
        \textbf{ITA:} Salute Mentale - \textbf{ENG:} Mental Health \\
        \hline
    \end{tabular}}
    \label{tab:psychotherapy_specialization}
\end{table}

In contrast, the \textbf{IMB-MCQA} dataset consists of structured multiple-choice questions derived from medical specialization exam simulators, providing a controlled environment for evaluating domain-specific knowledge (an example is shown in Table \ref{tab:dermatitis_mcqa}).

\begin{table}[t]
    \centering
    \caption{Example of a multiple-choice question from \textbf{IMB-MCQA}.}
    \scalebox{0.76}{
    \begin{tabular}{p{2.5cm} p{6cm}}
        \hline
        \multicolumn{2}{c}{\textbf{IMB-MCQA}} \\
        \hline
        \textbf{ID} & 121 \\ 
        \hline
        \textbf{Category} &  
        \textbf{ITA:} Dermatologia e venereologia 
        
        \textbf{ENG:} Dermatology and Venereology \\ 
        \hline
        \textbf{Question} &  
        \textbf{ITA:} Dermatite da contatto: quale delle affermazioni sottoriportate è corretta?  
        
        \textbf{ENG:} Dermatitis: which of the following statements is correct? \\ 
        \hline
        \textbf{Answer A} &  
        \textbf{ITA:} È una genodermatosi
        
        \textbf{ENG:} It is a genodermatosis \\ 
        \hline
        \textbf{Answer B} &  
        \textbf{ITA:} È più frequente negli individui di razza nera 
        
        \textbf{ENG:} It is more common in individuals of African descent \\ 
        \hline
        \textbf{Answer C} &  
        \textbf{ITA:} È causata spesso dall'uso di cosmetici 
        
        \textbf{ENG:} It is often caused by the use of cosmetics \\ 
        \hline
        \textbf{Answer D} &  
        \textbf{ITA:} Si realizza al 1° contatto con l'allergene 
        
        \textbf{ENG:} It occurs at the first contact with the allergen \\ 
        \hline
        \textbf{Answer E} &  
        \textbf{ITA:} Tutte le precedenti
        
        \textbf{ENG:} All of the above \\ 
        \hline
        \textbf{Percentage Correct} & 49\% \\ 
        \hline
        \textbf{Correct Answer} &  
        \textbf{ITA:} È causata spesso dall'uso di cosmetici  
        
        \textbf{ENG:} It is often caused by the use of cosmetics \\ 
        \hline
    \end{tabular}}
    \label{tab:dermatitis_mcqa}
\end{table}

\subsection{Data Collection}
The \textbf{IMB-QA} dataset was constructed by collecting questions and answers from two Italian medical forums: MedicItalia and Dica33. These public platforms facilitate interactions between users and certified healthcare professionals. The selection of these forums was guided by qualitative reliability criteria, including verification of medical credentials and assessment of response quality. The data extraction process was conducted through automated retrieval of publicly available information. To enhance compliance with GDPR requirements, an anonymization procedure was applied to remove Personally Identifiable Information (PII). However, we acknowledge that ensuring complete anonymization is inherently challenging, especially in medical contexts where indirect re-identification risks may persist. Future iterations of the dataset will incorporate additional validation steps to assess and improve the effectiveness of the anonymization process. The dataset covers a broad spectrum of common clinical conditions, supporting its medical representativeness. Each sample consists of the following components: A \textit{question} formulated by a user, representing a real medical concern and assigned to a specific medical category; An \textit{answer} provided by a certified healthcare professional, reformulated when necessary to improve clarity and coherence while ensuring the anonymization of personal data; Additional \textit{metadata}, including the \textit{corresponding medical category}, the \textit{macro-category}, and, where applicable, the \textit{URL} of the original source.  

The \textbf{IMB-MCQA} dataset, on the other hand, was constructed by collecting multiple-choice questions from Italian medical specialization exam simulator CompitoInClasse.org. Each sample consists of the following components:  A \textit{question} related to a specific clinical topic, selected from official simulators that provide access to past examination questions; The \textit{multiple-choice answers} associated with the question, including one correct answer validated by domain experts; The \textit{medical category} of the question, identifying the relevant medical field (e.g., physiology, cardiology, etc.); The \textit{percentage of correct answers}, calculated based on responses from a substantial number of candidates who have used the simulator, with a minimum response threshold to ensure reliability.

\subsection{Data preprocessing methods}
The \textbf{IMB-QA} dataset was built from Italian medical forums, collecting 782,644 patient questions and certified professional answers across 77 categories (up to July 2024), capturing real-world interactions.

The \textbf{IMB-MCQA} dataset was compiled from official Italian medical specialization exams through 2024 and includes 25,862 multiple-choice questions across ~60 clinical fields, each with 4–5 options. As typical with unstructured sources, both datasets had inconsistencies, redundancies, and PII. A multi-stage preprocessing pipeline improved their quality and NLP usability. Summary statistics are in Table \ref{tab:overall_stats}.

\subsubsection{Preprocessing for IMB-QA}
\paragraph{Data cleaning} Incomplete/truncated questions were removed, doctor signatures and timestamps stripped, and minor inconsistencies fixed, preserving meaning.

\paragraph{Text Normalization, Answer Reformulation, and Data Anonymization} These operations were carried out using Llama3-Med42-8B \cite{llmMed42}, a Large Language Model (LLM) specialized in the medical domain and adapted for multilingual tasks. The model underwent a \textit{prompt engineering} phase to enhance the clarity, coherence, and grammatical accuracy of the responses while preserving an adequate level of fidelity to medical information. User-submitted questions were retained in their original form to preserve the natural variability and authenticity of real-world patient inputs. In contrast, doctors' responses were reformulated according to three main criteria: (i) removal of redundancies and colloquial language, (ii) stylistic consistency across responses, and (iii) improved readability for more effective processing by NLP models. To address anonymization, we utilized Italian\_NER\_XXL \cite{ItalianNERXXL}, a NER model specifically trained in Italian. This model successfully identified PII, such as names of patients and doctors, cities, online resources, email addresses, healthcare facilities, and other identifiers that could enable re-identification. The identified PII underwent an anonymization procedure using the same LLM employed for reformulation, which preserved sentence semantics while substituting terms with generic medical context-appropriate alternatives. The effectiveness of anonymization was evaluated by calculating the percentage of PII --- detected using the same NER model as in the anonymization phase --- in the initial, reformulated, and anonymized responses on a subset of approximately 2163 responses equally selected from all medical categories in the dataset. Initially, 27\% of answers contained PII; reformulation reduced this to 7\%, and ultimately, anonymization decreased the presence of PII to just 1\%.

\begin{table}[t]
    \centering
    \caption{Overall statistics for \textbf{IMB-QA} and \textbf{IMB-MCQA}.}
    \scalebox{0.95}{
    \begin{tabular}{lrr}
        \hline
        \textbf{Statistic} & \textbf{IMB-QA} & \textbf{IMB-MCQA}    \\ 
        \hline
        \textbf{\# Questions and Answers} & 782,644 & 25,862        \\ 
        \textbf{\# Categories} & 77 & 60                            \\ 
        \textbf{Last Update} & July 2024 & July 2024                \\ 
        \hline
        \textbf{Tot. Answer Tokens} & 40,370,381 & 9,321             \\ 
        \textbf{Unique Answer Vocab.} & 154,837 & 1,234               \\ 
        \textbf{Tot. Question Tokens} & 137,129,435 & 282,239       \\ 
        \textbf{Unique Question Vocab.} & 1,397,929 & 19,214        \\ 
        \textbf{Unique Total Vocab.} & 1,552,766 & 20,448              \\ 
        \hline
        \textbf{Avg. Answer Length} & 352.05 & 9.3                  \\
        \textbf{Max. Answer Length} & 9,817 & 21                   \\ 
        \textbf{Avg. Question Length} & 1,056.77 & 10.91            \\ 
        \textbf{Max. Question Length} & 13,390 & 124                \\ 
        \hline
    \end{tabular}}
    \label{tab:overall_stats}
\end{table}

\begin{figure*}[t]
    \centering \includegraphics[width=\textwidth]{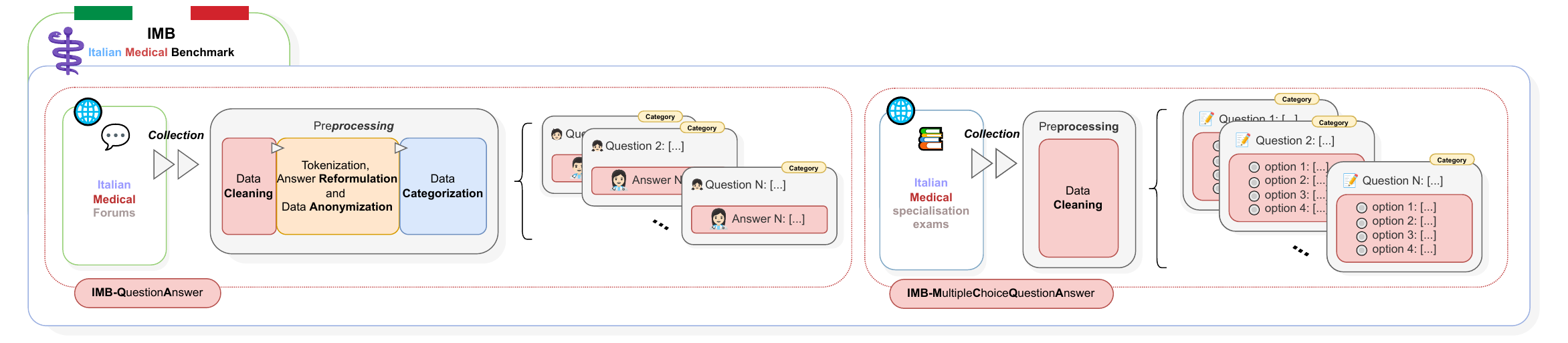}
    \caption{Workflow for the construction of the Italian Medical Benchmark (IMB), consisting of open-ended question-answer pairs (IMB-QA) and multiple-choice question-answer assessments (IMB-MCQA).}
    \label{fig:Workflow}
\end{figure*}

\paragraph{Data Categorization} To group questions into broader semantic fields, unsupervised topic modeling via BERTopic \cite{Berttopic} was applied. Sentence embeddings were generated with "paraphrase-multilingual-MiniLM-L12-v2" \cite{reimers-2019-sentence-bert}, reduced via UMAP \cite{UMAP}, and clustered using HDBSCAN \cite{HDBSCAN}. This enabled flexible, interpretable macro-categorization without enforcing rigid class definitions. Final groupings are reported in Table \ref{tab:category_stats}.

\begin{table}[t]
    \centering
    \caption{Macro-categories and number of related questions in \textbf{IMB-QA}.}
    \scalebox{0.95}{
    \begin{tabular}{p{5cm} c}
        \hline
        \textbf{Category} & \textbf{N.o Questions}       \\ 
        \hline
        Urology, andrology and male health & 110,052               \\
        Gastroenterology and digestive health & 104,449            \\
        Mental health & 103,893                                    \\
        General Medicine and General Surgery & 87,789              \\
        Ophthalmology, otolaryngology, dentistry and pneumology & 83,710            \\
        Cardiology, circulatory system and hematology & 81,232     \\
        Gynecology and female health & 65,792                      \\
        Orthopedics and musculoskeletal system & 50,283            \\
        Dermatology, allergies and aesthetics & 49,288             \\
        Neurology & 46,704                                         \\
        \hline
    \end{tabular}}
    \label{tab:category_stats}
\end{table}

\subsubsection{Preprocessing for IMB-MCQA}
As this dataset was already in a clean, structured exam format, preprocessing mainly involved organizing entries and ensuring consistent formatting. No major cleaning or reformulation was necessary. The workflow is summarized in Figure \ref{fig:Workflow}.

\subsection{Data Analysis}

\subsubsection{Diversity of Questions}
Clinical medicine covers a broad range of topics, reflected in the question types within the \textbf{IMB} dataset. To assess this variety, a qualitative analysis was conducted on a random sample of 102 questions from \textbf{IMB-QA} and \textbf{IMB-MCQA}. Given the complexity of accurately classifying questions as \textbf{fact-based} or \textbf{case-based} through automated methods, manual categorization was chosen. \textbf{Fact-based} questions focus on specific medical knowledge and clear reasoning, such as “Which condition is linked to persistent fatigue?”. \textbf{Case-based} questions, instead, present a patient's symptoms or medical background, requiring multi-step reasoning for diagnosis, treatment decisions, or prognosis, such as assessing a patient with chest pain. The analysis indicates that \textbf{IMB-QA} is predominantly composed of \textbf{case-based} questions, where patients describe symptoms and seek medical guidance, requiring models to perform complex reasoning. Although \textbf{IMB-MCQA} mainly consists of \textbf{fact-based} questions, as it evaluates medical knowledge for specialization exams, it also includes a considerable number of \textbf{case-based} inquiries. This dual function highlights the dataset's role in assessing both factual knowledge and clinical decision-making, with \textbf{IMB-QA} emphasizing patient narratives and \textbf{IMB-MCQA} blending factual recall with clinical reasoning.

\subsubsection{Need for Domain-Specific Expertise} \label{sec:complexity_score}
\begin{figure}[t]
    \centering \includegraphics[width=0.48\textwidth]{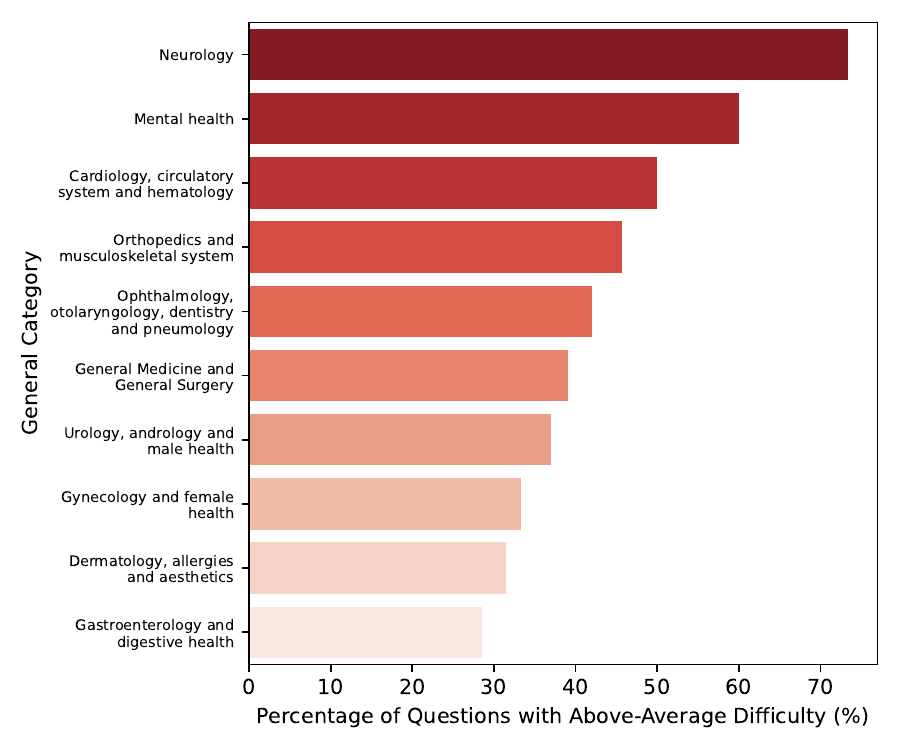}
    \caption{Percentage of questions with above-average difficulty by macro-category in \textbf{IMB-QA}. The score refers to the percentage of questions in each category that were classified as above-average in difficulty, based on our difficulty index}
    \label{fig:hypotesis2}
\end{figure}

To evaluate the datasets' complexity, we assessed question difficulty. In \textbf{IMB-QA}, a sample of 2,500 questions was analyzed using a difficulty index based on length, terminology, and syntax. 39.24\% were above-average in difficulty, with Neurology exceeding 70\%, indicating high specialization demands (Figure \ref{fig:hypotesis2}).

\begin{figure}[t]
    \centering \includegraphics[width=0.48\textwidth]{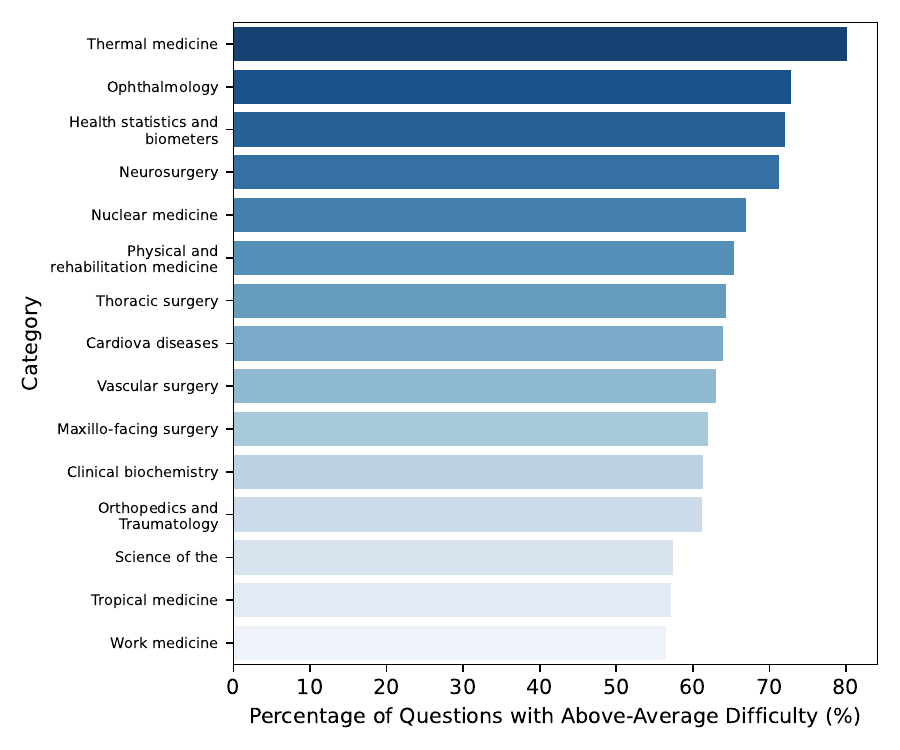}
    \caption{Percentage of questions with above-average difficulty by category in \textbf{IMB-MCQA}. The score refers to the percentage of questions in each category that were classified as above-average in difficulty, based on our difficulty index }
    \label{fig:hypotesis21}
\end{figure}

In \textbf{IMB-MCQA}, difficulty was estimated from participant accuracy. Categories like "Thermal Medicine" (80.12\%), "Ophthalmology" (72.86\%), "Neurosurgery" (71.30\%), and "Nuclear Medicine" (66.95\%) showed high complexity (Figure \ref{fig:hypotesis21}).

These results confirm that both datasets require advanced clinical knowledge, making them valuable for training models in specialized medical reasoning.

\subsubsection{Diversity of Categories}
\begin{figure*}[!t]
    \centering
    \begin{minipage}{0.5\textwidth}
        \centering
        \includegraphics[width=\textwidth]{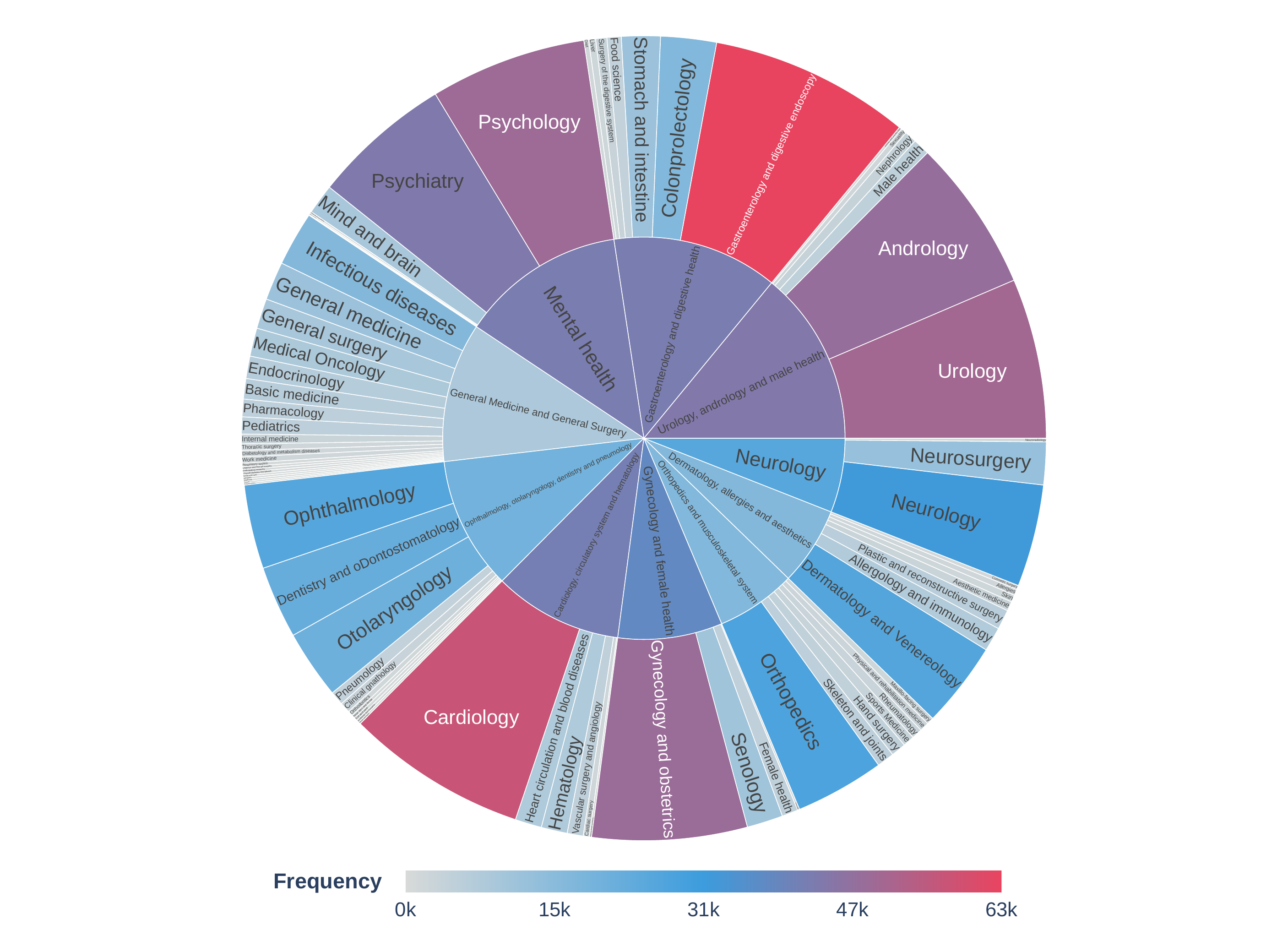}
        \caption{Distribution of macro-categories in \textbf{IMB-QA}.}
        \label{fig:Ventaglio1}
    \end{minipage}
    \hfill
    \begin{minipage}{0.49\textwidth}
        \centering
        \includegraphics[width=\textwidth]{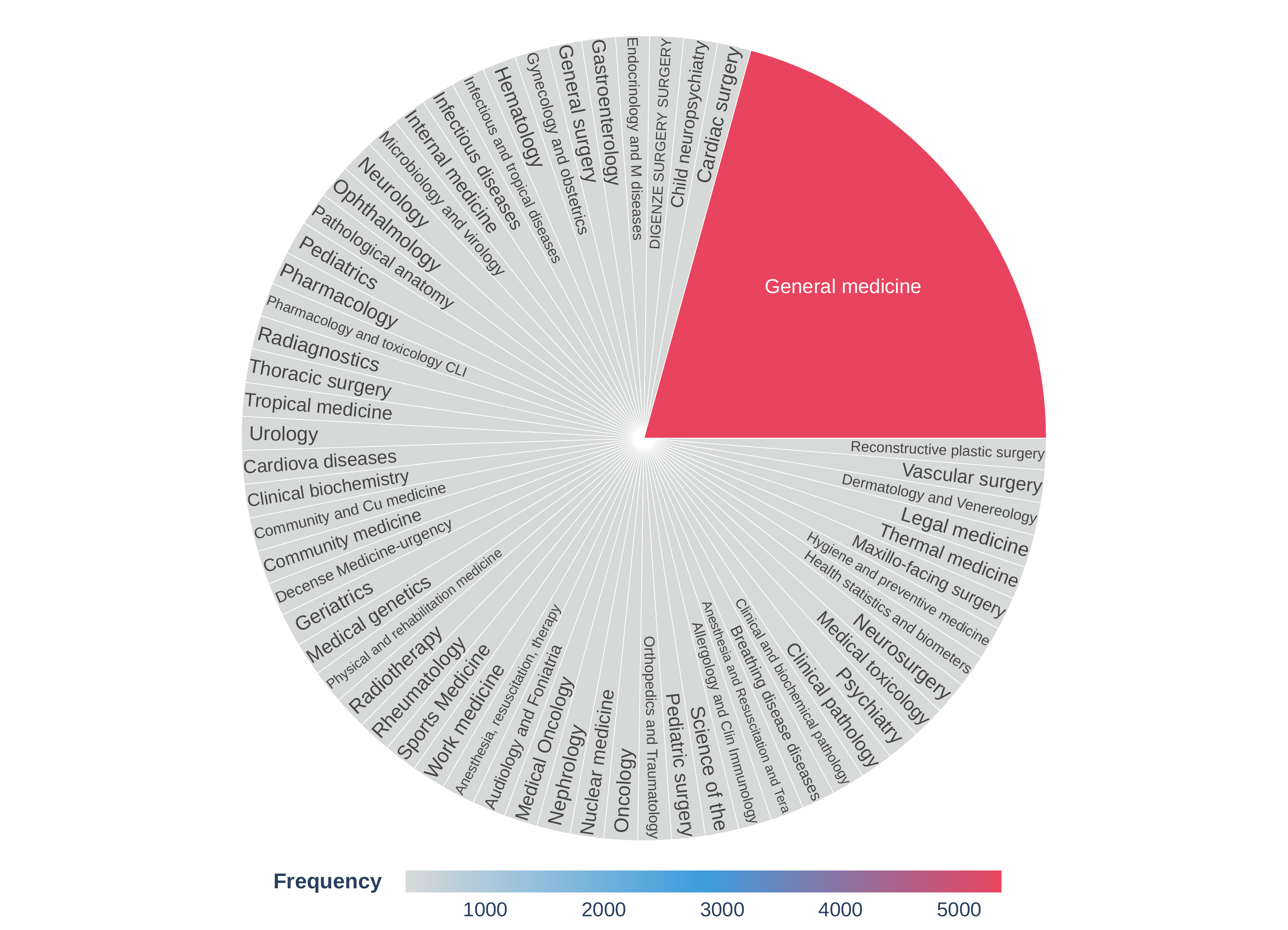}
        \caption{Distribution of categories in \textbf{IMB-MCQA}.}
        \label{fig:Ventaglio2}
    \end{minipage}
\end{figure*}

The \textbf{IMB} dataset shows uneven category distribution, affecting model performance across specialties. \textbf{IMB-QA} (Figure \ref{fig:Ventaglio1}) overrepresents areas like "Gastroenterology", "Cardiology", and "Urology", while fields like "Sleep Medicine" and "Pediatric Surgery" are underrepresented. This may lead to imbalanced model capabilities. \textbf{IMB-MCQA} (Figure \ref{fig:Ventaglio2}) shows a more uniform distribution, with most categories having $\sim$350 questions, except "General Medicine" ($\sim$5,000), reducing but not eliminating coverage gaps in niche fields.

\subsubsection{Presence of Information Noise and Ambiguity in Responses}
Challenges in the \textbf{IMB} dataset include noise and ambiguity. In \textbf{IMB-QA}, informal forum responses often contain contextual or generic advice, sometimes prioritizing in-person consultation over definitive answers. These traits, while realistic, introduce variability. Preprocessing helped filter irrelevant elements and standardize responses. In \textbf{IMB-MCQA}, ambiguity stems from distractors designed to assess reasoning, with some questions allowing multiple valid interpretations. Such complexity enhances the dataset’s value in training models to manage uncertainty and emulate clinical decision-making.

\section{Applications} \label{sec: applications}

\subsection{Benchmarking Large Language Models}
Evaluating LLMs on domain-specific datasets is essential to measure their suitability for fields like medicine, where precise understanding is required \cite{liu2023nips}. Despite advancements in general-purpose knowledge, performance in non-English clinical contexts remains limited \cite{jin2024www}. \textbf{IMB-QA} and \textbf{IMB-MCQA} enable benchmarking in Italian for both open-ended and multiple-choice medical QA, capturing language-specific features, technical terminology, and clinical nuances.

\begin{table}[t]
    \centering
    \caption{Language models benchmarked in our experiments.}
    \scalebox{0.90}{
    \begin{tabular}{l ccc}
        \toprule
        \textbf{Model} & \textbf{Size} & \textbf{Fine-tuned} & \textbf{Language} \\
        \midrule
        Mistral-7B-Instruct-v0.3 & 7B & No & English \\
        LLaMa-3.1-70B-Instruct & 70B & No & English \\
        LLaMa-3.1-8B-Instruct & 8B & No & English \\
        LLaMa-3.2-3B-Instruct & 3B & No & English \\
        Gemma-2-9b-it & 9B & No & English \\
        BioMistral-7B & 7B & Yes & English \\
        Bio-Medical-Llama-3-8B & 8B & Yes & English \\
        Maestrale-Chat-v0.4 & 7B & Yes & Italian \\
        LLaMAntino 3-8B & 8B & Yes & Italian \\
        Velvet-14B & 14B & No & Italian \\
        \bottomrule
    \end{tabular}}
    \label{tab:benchmark_models}
\end{table}

We evaluate open-ended QA using BERTScore \cite{zhang2020iclr} with the multilingual model \texttt{bert-base-multilingual-cased}, chosen for its cross-lingual semantic similarity capabilities and its widespread adoption in multilingual NLP benchmarks. For MCQA tasks, we report standard accuracy. This dual evaluation highlights LLM strengths and limitations in Italian clinical applications.

\subsection{Medical Question Answering}
Medical QA demands models that handle informal, complex queries without hallucinating \cite{zhang2017cMedQA, DBLP:journals/corr/abs-2401-11389}. We apply \textbf{Retrieval-Augmented Generation} (RAG) using a separate knowledge base of 100k anonymized \textit{IMB-QA} answers, explicitly excluding evaluation samples to avoid data leakage. Relevant contexts are retrieved via dense embeddings generated with \texttt{all-MiniLM-L6-v2}\footnote{\url{https://huggingface.co/sentence-transformers/all-MiniLM-L6-v2}} and indexed using FAISS \cite{douze2024faiss}. We retrieve the top-5 most similar passages, which are then prepended to the query. This ensures factual grounding while maintaining separation between retrieved context and target answers. Although we did not perform a separate retriever evaluation, the overall gain in BERTScore (Table~\ref{tab:rag_improvements}) confirms the added value of retrieval. The process is formalized as:
\begin{equation}
A = \text{LLM}(Q, R(Q, D))
\end{equation}
where \( Q \) is the query, \( D \) the dataset, and \( R \) the retrieval function. Table~\ref{tab:rag_improvements} shows RAG improves BERTScore Precision across all categories.

\begin{table}[t] 
    \centering
    \caption{BERTScore Precision: gemma-2-9b-it with and without RAG on \textbf{IMB-QA}.}
    \scalebox{0.90}{
    \begin{tabular}{l ccc} 
        \toprule 
        \textbf{Category} & \textbf{w/o RAG} & \textbf{RAG} & \textbf{$\Delta \%$} \\ 
        \midrule 
        Cardiology, hematology & 0.632 & \textbf{0.672} & 6.33\% \\ 
        Dermatology, aesthetics & 0.636 & \textbf{0.678} & 6.60\% \\ 
        Gastroenterology & 0.638 & \textbf{0.679} & 6.42\% \\ 
        General medicine & 0.636 & \textbf{0.674} & 5.97\% \\ 
        Gynecology & 0.630 & \textbf{0.671} & 6.51\% \\ 
        Mental health & 0.636 & \textbf{0.677} & 6.45\% \\ 
        ENT, ophthalmology & 0.647 & \textbf{0.685} & 5.87\% \\ 
        Orthopedics & 0.628 & \textbf{0.669} & 6.52\% \\ 
        Urology, andrology & 0.638 & \textbf{0.679} & 6.42\% \\ 
        Neurology & 0.653 & \textbf{0.706} & 8.12\% \\ 
        \bottomrule 
    \end{tabular}}
    \label{tab:rag_improvements}
\end{table}

\subsection{Fine-tuning}
Fine-tuning improves domain alignment for LLMs, especially in non-English medical contexts \cite{tran2024jamia, DBLP:journals/corr/abs-2408-13833}. Using \textbf{IMB-QA}, we fine-tune Small Language Models (SLMs) like Llama-3.2-1B, Gemma-2-2b-it, and Qwen2.5-1.5B \cite{van2024survey}, leveraging [CLS]/[SEP] token strategies, cross-entropy loss, and Curriculum Learning \cite{curriculum10.1145/1553374.1553380} via the Unsloth \cite{unsloth} library. This approach aims to enhance output accuracy and reduce hallucinations while ensuring efficient deployment in clinical environments.
Although formal hallucination metrics are not reported, results in Table~\ref{tab:ft_vs_noft} show that fine-tuning on \textbf{IMB-QA} leads to modest improvements across several metrics, particularly in BERTScore and BLEU. Gains are model-dependent and not uniform across all scores: for instance, METEOR slightly decreases in some cases. Nonetheless, the overall trend supports the effectiveness of task-specific adaptation in improving answer quality in Italian medical QA.

\section{Experiments} \label{sec: experiments}

\subsection{Experimental Setup}
Experiments were conducted on Google Colab Pro using an NVIDIA T4 GPU and Intel Xeon CPU. Due to hardware constraints, the evaluation focused on the most complex categories, as defined in Section \ref{sec:complexity_score}. For \textbf{IMB-QA}, $\sim$2,000 instances were sampled per category, except for the "Neurology" category, which includes only 998 instances. In the case of \textbf{IMB-MCQA}, the full set of instances for each category was used. Models were implemented with Hugging Face Transformers and fine-tuned using the Unsloth library, leveraging mixed precision (fp16) to optimize memory and convergence speed. Each model was fine-tuned for 6 epochs using the Cross Entropy loss function and a fixed learning rate of $2.97e^{-4}$.

\subsection{Benchmarking LLMs \& SLMs Results}
\begin{figure*}[t]
    \centering
    \begin{minipage}{0.49\textwidth}
        \centering
        \includegraphics[width=.99\textwidth]{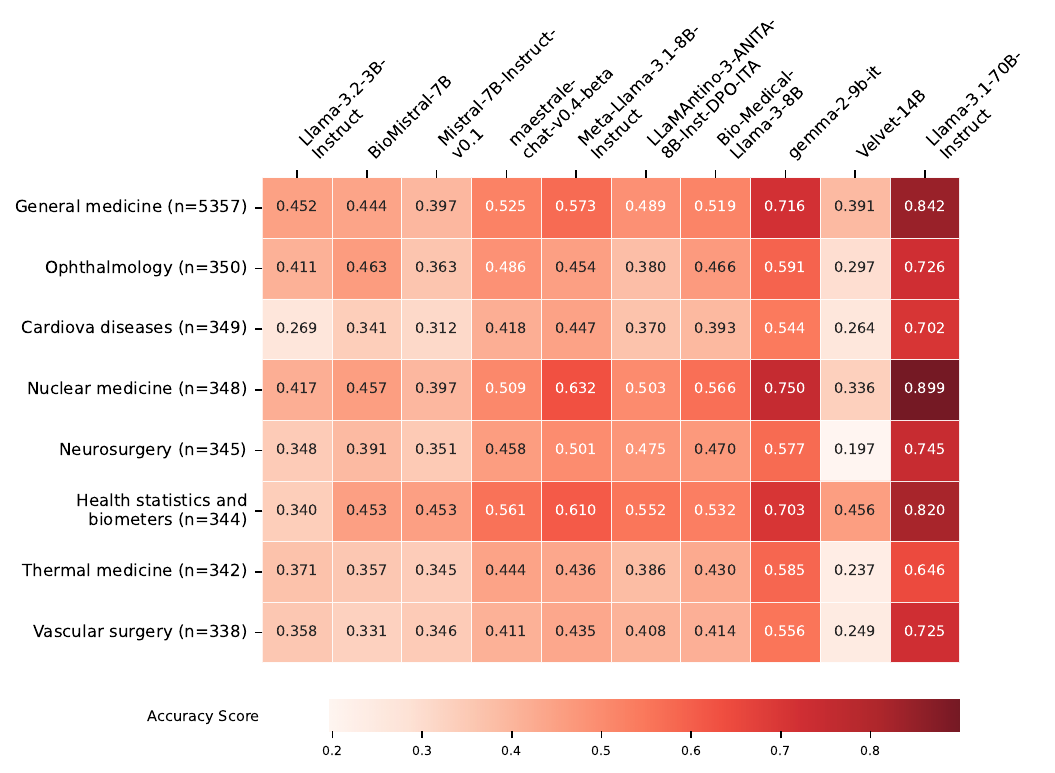}
        \caption{LLM benchmark on \textbf{IMB-MCQA}.}
        \label{fig:heatmapLLM_AccuracyScore}
    \end{minipage}
    \hfill
    \begin{minipage}{0.49\textwidth}
        \centering
        \includegraphics[width=\textwidth]{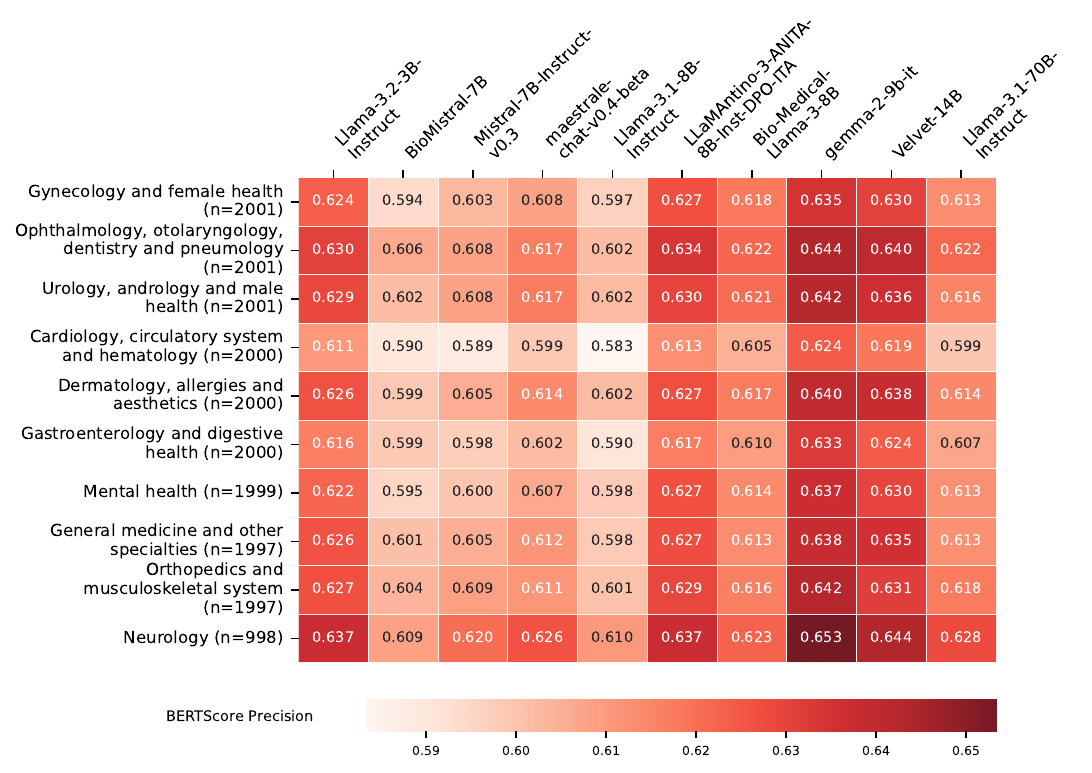}
        \caption{LLM benchmark on \textbf{IMB-QA}.}
        \label{fig:heatmapLLM_BERTScore}
    \end{minipage}
\end{figure*}

\textbf{IMB-MCQA} offers a robust benchmark for clinical QA in multiple-choice format, evaluated using accuracy. As shown in Figure~\ref{fig:heatmapLLM_AccuracyScore}, models with more than 8B parameters achieve nearly 85\% accuracy, outperforming smaller models, which struggle with domain-specific reasoning. These trends align with prior analyses of category difficulty, where questions involving underrepresented or cognitively complex fields proved more challenging even for advanced LLMs.

\subsection{Medical QA Results}
\textbf{IMB-QA} allows assessment of open-ended medical QA, where semantic accuracy is paramount. In Figure~\ref{fig:heatmapLLM_BERTScore}, \texttt{gemma-2-9b-it} outperforms larger models, likely due to its multilingual training. Despite its smaller size, it achieves competitive BERTScore Precision (up to 0.638), suggesting high semantic alignment. This metric is more informative than fluency-based ones in clinical settings, where accurate, relevant answers are crucial.

\subsection{Fine-tuning SLMs Results}
We fine-tuned several SLMs, including \texttt{Llama-3.2-3B}, on \textbf{IMB-QA} using an 80/20 train/eval split and leveraging Unsloth library. As shown in Table~\ref{tab:ft_vs_noft}, fine-tuned models generally showed modest improvements over base versions, although gains varied across metrics and models, with some showing performance drops in specific scores such as METEOR. This confirms that task adaptation improves answer quality and contextual understanding, even for compact models, making them well-suited for clinical applications.

\begin{table*}[t]
    \centering
    \caption{Comparison between fine-tuned and non-fine-tuned models on \textbf{IMB-QA}.}
    \scalebox{0.75}{
    \begin{tabular}{l ccccccccc}
        \hline
        \textbf{Model} & \textbf{Fine-Tuned} & \textbf{ROUGE-1} & \textbf{ROUGE-2} & \textbf{ROUGE-L} & \textbf{BLEU} & \textbf{METEOR} & \textbf{BERTScore P} & \textbf{BERTScore R} & \textbf{BERTScore F1} \\
        \hline
        \multirow{2}{*}{Llama-3.2-1B-Instruct} & Yes & \cellcolor{green!25} 0.2857 & \cellcolor{green!25} 0.0572 & \cellcolor{green!25} 0.1998 & \cellcolor{green!25} 0.0309 & 0.1682 & \cellcolor{green!25} 0.7107 & \cellcolor{green!25} 0.6860 & \cellcolor{green!25} 0.6976 \\
        & No & 0.2315 & 0.0445 & 0.1552 & 0.0148 & \cellcolor{green!25} 0.2137 & 0.6186 & 0.6680 & 0.6423 \\ \midrule
        \multirow{2}{*}{gemma-2-2b-it} & Yes & 0.2673 & \cellcolor{green!25} 0.0586 & 0.1890 & \cellcolor{green!25} 0.0336 & 0.1617 & \cellcolor{green!25} 0.7098 & 0.6775 & \cellcolor{green!25} 0.6926 \\
        & No & \cellcolor{green!25} 0.2932 & 0.0511 & \cellcolor{green!25} 0.1918 & 0.0228 & \cellcolor{green!25} 0.2055 & 0.6783 & \cellcolor{green!25} 0.6870 & 0.6821 \\ \midrule
        \multirow{2}{*}{Llama-3.2-3B-Instruct} & Yes & \cellcolor{green!25} 0.2994 & \cellcolor{green!25} 0.0642 & \cellcolor{green!25} 0.1995 & \cellcolor{green!25} 0.0424 & 0.1952 & \cellcolor{green!25} 0.7031 & \cellcolor{green!25} 0.6924 & \cellcolor{green!25} 0.6972 \\
        & No & 0.2523 & 0.0509 & 0.1607 & 0.0213 & \cellcolor{green!25} 0.2310 & 0.6332 & 0.6830 & 0.6569 \\ \midrule
        \multirow{2}{*}{Qwen2.5-1.5B-Instruct} & Yes & \cellcolor{green!25} 0.2628 & \cellcolor{green!25} 0.0438 & \cellcolor{green!25} 0.1761 & \cellcolor{green!25} 0.0201 & \cellcolor{green!25} 0.1571 & \cellcolor{green!25} 0.7049 & \cellcolor{green!25} 0.6859 & \cellcolor{green!25} 0.6948 \\
        & No & 0.1141 & 0.0180 & 0.0756 & 0.0103 & 0.1283 & 0.6021 & 0.6617 & 0.6302 \\
        \hline
    \end{tabular}}
    \label{tab:ft_vs_noft}
\end{table*}

\section{Conclusion \& Future Work} \label{sec: conclusion}
In this work, we introduced IMB, the first Italian dataset for medical question-answering, which includes both open-ended (QA) and multiple-choice (MCQA) questions. The dataset, sourced from medical forums and exam simulators, provides a valuable resource for the development of advanced NLP models. Our qualitative and quantitative analysis highlighted a diverse range of medical specialties, while also revealing challenges related to question difficulty and clinical complexity. Initial experiments with state-of-the-art language models demonstrated that these models struggle with clinically complex Italian questions but perform relatively well on multiple-choice questions. Future work will focus on expanding the dataset by incorporating additional medical specialties and languages (such as English), improving category balancing, and implementing advanced filtering techniques to reduce informational noise. Furthermore, we will explore strategies for adapting language models to improve their ability to understand and reason effectively about medical content.

\paragraph{Limitations}
\textbf{IMB} has several limitations, including an imbalance in specialty representation. Fields such as "Gastroenterology" and "Cardiology" are overrepresented, while others, such as "Sleep Medicine" and "Pediatric Surgery", have limited coverage. This imbalance may affect model generalization. We will address this issue through data balancing techniques, such as oversampling and weighted training strategies. Another limitation arises from informational noise, as the questions were automatically collected from public sources, which may include irrelevant or ambiguous details. We plan to tackle this challenge by employing semantic filtering and human verification methods. Additionally, ambiguity in responses, particularly in the \textbf{IMB-MCQA} dataset, poses a challenge, which we aim to overcome through disambiguation techniques and more precise annotation strategies.

\paragraph{Ethical and Legal Considerations}
Our dataset has been developed using content sourced information from publicly accessible Italian medical sites (MedicItalia, Dica33) as well as a medical exam simulator (CompitoInClasse.org). The dataset is intended exclusively for academic research with non-commercial objectives, adhering to legal guidelines regarding GDPR compliance, data anonymization, and research-related copyright exemptions as outlined in Italian and EU legislation. To mitigate any legal and ethical challenges, and based on consultations with legal experts, we implemented several measures: \textbf{(1) Anonymization:} All identifying details (e.g. names, contact details, emails) were removed or altered with the help of automated scripts and LLM-supported redaction, conforming to GDPR's tenets of data minimization and protection. \textbf{(2) Textual Transformation:} While we provide links to the original source of each data sample, the raw questions and answers were linguistically restructured and polished, involving grammatical adjustments, simplification, and content refinement with the aid of LLMs and manual oversight. \textbf{(3) Scientific Scope:} This data serves strictly educational, illustrative, and scientific purposes as permitted under Article 89 of the GDPR and Article 70 of the Italian Copyright Law, which allows non-commercial research data usage under specified conditions. For this reason, the dataset is distributed under a \textit{Creative Commons Attribution-NonCommercial-NoDerivatives 4.0 (CC BY-NC-ND 4.0)} license. This license strictly restricts usage to non-commercial research, prohibits redistribution of altered versions, and mandates proper author attribution.

\begin{acknowledgments}
   This work was conducted with the financial support of (1) the PNRR MUR project PE0000013-FAIR and (2) the Italian ministry of economic development, via the ICARUS (Intelligent Contract Automation for Rethinking User Services) project (CUP: B69J23000270005).
\end{acknowledgments}

\bibliography{sample-ceur}

\begin{thebibliography}{41}
\expandafter\ifx\csname natexlab\endcsname\relax\def\natexlab#1{#1}\fi
\providecommand{\url}[1]{\texttt{#1}}
\providecommand{\href}[2]{#2}
\providecommand{\path}[1]{#1}
\providecommand{\DOIprefix}{doi:}
\providecommand{\ArXivprefix}{arXiv:}
\providecommand{\URLprefix}{URL: }
\providecommand{\Pubmedprefix}{pmid:}
\providecommand{\doi}[1]{\href{http://dx.doi.org/#1}{\path{#1}}}
\providecommand{\Pubmed}[1]{\href{pmid:#1}{\path{#1}}}
\providecommand{\bibinfo}[2]{#2}
\ifx\xfnm\relax \def\xfnm[#1]{\unskip,\space#1}\fi
\bibitem[{Devlin et~al.(2019)Devlin, Chang, Lee, and Toutanova}]{devlin-etal-2019-bert}
\bibinfo{author}{J.~Devlin}, \bibinfo{author}{M.-W. Chang}, \bibinfo{author}{K.~Lee}, \bibinfo{author}{K.~Toutanova},
\newblock \bibinfo{title}{{BERT}: Pre-training of deep bidirectional transformers for language understanding},
\newblock in: \bibinfo{editor}{J.~Burstein}, \bibinfo{editor}{C.~Doran}, \bibinfo{editor}{T.~Solorio} (Eds.), \bibinfo{booktitle}{Proceedings of the 2019 Conference of the North {A}merican Chapter of the Association for Computational Linguistics: Human Language Technologies, Volume 1 (Long and Short Papers)}, \bibinfo{publisher}{Association for Computational Linguistics}, \bibinfo{address}{Minneapolis, Minnesota}, \bibinfo{year}{2019}, pp. \bibinfo{pages}{4171--4186}. \URLprefix \url{https://aclanthology.org/N19-1423/}. \DOIprefix\doi{10.18653/v1/N19-1423}.
\bibitem[{Liu et~al.(2019)Liu, Ott, Goyal, Du, Joshi, Chen, Levy, Lewis, Zettlemoyer, and Stoyanov}]{DBLP:journals/corr/abs-1907-11692}
\bibinfo{author}{Y.~Liu}, \bibinfo{author}{M.~Ott}, \bibinfo{author}{N.~Goyal}, \bibinfo{author}{J.~Du}, \bibinfo{author}{M.~Joshi}, \bibinfo{author}{D.~Chen}, \bibinfo{author}{O.~Levy}, \bibinfo{author}{M.~Lewis}, \bibinfo{author}{L.~Zettlemoyer}, \bibinfo{author}{V.~Stoyanov},
\newblock \bibinfo{title}{Roberta: {A} robustly optimized {BERT} pretraining approach},
\newblock \bibinfo{journal}{CoRR} \bibinfo{volume}{abs/1907.11692} (\bibinfo{year}{2019}) \bibinfo{pages}{--}. \URLprefix \url{http://arxiv.org/abs/1907.11692}. \href{http://arxiv.org/abs/1907.11692}{{\tt arXiv:1907.11692}}.
\bibitem[{Lee et~al.(2020)Lee, Yoon, Kim, Kim, Kim, So, and Kang}]{DBLP:journals/bioinformatics/LeeYKKKSK20}
\bibinfo{author}{J.~Lee}, \bibinfo{author}{W.~Yoon}, \bibinfo{author}{S.~Kim}, \bibinfo{author}{D.~Kim}, \bibinfo{author}{S.~Kim}, \bibinfo{author}{C.~H. So}, \bibinfo{author}{J.~Kang},
\newblock \bibinfo{title}{Biobert: a pre-trained biomedical language representation model for biomedical text mining},
\newblock \bibinfo{journal}{Bioinform.} \bibinfo{volume}{36} (\bibinfo{year}{2020}) \bibinfo{pages}{1234--1240}. \URLprefix \url{https://doi.org/10.1093/bioinformatics/btz682}. \DOIprefix\doi{10.1093/BIOINFORMATICS/BTZ682}.
\bibitem[{Rajpurkar et~al.(2016)Rajpurkar, Zhang, Lopyrev, and Liang}]{rajpurkar-etal-2016-squad}
\bibinfo{author}{P.~Rajpurkar}, \bibinfo{author}{J.~Zhang}, \bibinfo{author}{K.~Lopyrev}, \bibinfo{author}{P.~Liang},
\newblock \bibinfo{title}{{SQ}u{AD}: 100,000+ questions for machine comprehension of text},
\newblock in: \bibinfo{editor}{J.~Su}, \bibinfo{editor}{K.~Duh}, \bibinfo{editor}{X.~Carreras} (Eds.), \bibinfo{booktitle}{Proceedings of the 2016 Conference on Empirical Methods in Natural Language Processing}, \bibinfo{publisher}{Association for Computational Linguistics}, \bibinfo{address}{Austin, Texas}, \bibinfo{year}{2016}, pp. \bibinfo{pages}{2383--2392}. \URLprefix \url{https://aclanthology.org/D16-1264/}. \DOIprefix\doi{10.18653/v1/D16-1264}.
\bibitem[{Yang et~al.(2019)Yang, Dai, Yang, Carbonell, Salakhutdinov, and Le}]{DBLP:conf/nips/YangDYCSL19}
\bibinfo{author}{Z.~Yang}, \bibinfo{author}{Z.~Dai}, \bibinfo{author}{Y.~Yang}, \bibinfo{author}{J.~G. Carbonell}, \bibinfo{author}{R.~Salakhutdinov}, \bibinfo{author}{Q.~V. Le},
\newblock \bibinfo{title}{Xlnet: Generalized autoregressive pretraining for language understanding},
\newblock in: \bibinfo{editor}{H.~M. Wallach}, \bibinfo{editor}{H.~Larochelle}, \bibinfo{editor}{A.~Beygelzimer}, \bibinfo{editor}{F.~d'Alch{\'{e}}{-}Buc}, \bibinfo{editor}{E.~B. Fox}, \bibinfo{editor}{R.~Garnett} (Eds.), \bibinfo{booktitle}{Advances in Neural Information Processing Systems 32: Annual Conference on Neural Information Processing Systems 2019, NeurIPS 2019, December 8-14, 2019, Vancouver, BC, Canada}, \bibinfo{publisher}{NeurIPS}, \bibinfo{address}{Vancouver, BC, Canada}, \bibinfo{year}{2019}, pp. \bibinfo{pages}{5754--5764}. \URLprefix \url{https://proceedings.neurips.cc/paper/2019/hash/dc6a7e655d7e5840e66733e9ee67cc69-Abstract.html}.
\bibitem[{Kwiatkowski et~al.(2019)Kwiatkowski, Palomaki, Redfield, Collins, Parikh, Alberti, Epstein, Polosukhin, Devlin, Lee, Toutanova, Jones, Kelcey, Chang, Dai, Uszkoreit, Le, and Petrov}]{kwiatkowski-etal-2019-natural}
\bibinfo{author}{T.~Kwiatkowski}, \bibinfo{author}{J.~Palomaki}, \bibinfo{author}{O.~Redfield}, \bibinfo{author}{M.~Collins}, \bibinfo{author}{A.~Parikh}, \bibinfo{author}{C.~Alberti}, \bibinfo{author}{D.~Epstein}, \bibinfo{author}{I.~Polosukhin}, \bibinfo{author}{J.~Devlin}, \bibinfo{author}{K.~Lee}, \bibinfo{author}{K.~Toutanova}, \bibinfo{author}{L.~Jones}, \bibinfo{author}{M.~Kelcey}, \bibinfo{author}{M.-W. Chang}, \bibinfo{author}{A.~M. Dai}, \bibinfo{author}{J.~Uszkoreit}, \bibinfo{author}{Q.~Le}, \bibinfo{author}{S.~Petrov},
\newblock \bibinfo{title}{Natural questions: A benchmark for question answering research},
\newblock \bibinfo{journal}{Transactions of the Association for Computational Linguistics} \bibinfo{volume}{7} (\bibinfo{year}{2019}) \bibinfo{pages}{452--466}. \URLprefix \url{https://aclanthology.org/Q19-1026/}. \DOIprefix\doi{10.1162/tacl_a_00276}.
\bibitem[{Tsatsaronis et~al.(2015)Tsatsaronis, Balikas, Malakasiotis, Partalas, Zschunke, Alvers, Weissenborn, Krithara, Petridis, Polychronopoulos, Almirantis, Pavlopoulos, Baskiotis, Gallinari, Arti{\`{e}}res, Ngomo, Heino, Gaussier, Barrio{-}Alvers, Schroeder, Androutsopoulos, and Paliouras}]{DBLP:journals/bmcbi/TsatsaronisBMPZ15}
\bibinfo{author}{G.~Tsatsaronis}, \bibinfo{author}{G.~Balikas}, \bibinfo{author}{P.~Malakasiotis}, \bibinfo{author}{I.~Partalas}, \bibinfo{author}{M.~Zschunke}, \bibinfo{author}{M.~R. Alvers}, \bibinfo{author}{D.~Weissenborn}, \bibinfo{author}{A.~Krithara}, \bibinfo{author}{S.~Petridis}, \bibinfo{author}{D.~Polychronopoulos}, \bibinfo{author}{Y.~Almirantis}, \bibinfo{author}{J.~Pavlopoulos}, \bibinfo{author}{N.~Baskiotis}, \bibinfo{author}{P.~Gallinari}, \bibinfo{author}{T.~Arti{\`{e}}res}, \bibinfo{author}{A.~N. Ngomo}, \bibinfo{author}{N.~Heino}, \bibinfo{author}{{\'{E}}.~Gaussier}, \bibinfo{author}{L.~Barrio{-}Alvers}, \bibinfo{author}{M.~Schroeder}, \bibinfo{author}{I.~Androutsopoulos}, \bibinfo{author}{G.~Paliouras},
\newblock \bibinfo{title}{An overview of the {BIOASQ} large-scale biomedical semantic indexing and question answering competition},
\newblock \bibinfo{journal}{{BMC} Bioinform.} \bibinfo{volume}{16} (\bibinfo{year}{2015}) \bibinfo{pages}{138:1--138:28}. \URLprefix \url{https://doi.org/10.1186/s12859-015-0564-6}. \DOIprefix\doi{10.1186/S12859-015-0564-6}.
\bibitem[{Wang et~al.(2024)Wang, Huang, Jackson, and Gao}]{DBLP:journals/tacl/WangHJ024}
\bibinfo{author}{D.~Wang}, \bibinfo{author}{Q.~Huang}, \bibinfo{author}{M.~Jackson}, \bibinfo{author}{J.~Gao},
\newblock \bibinfo{title}{Retrieve what you need: {A} mutual learning framework for open-domain question answering},
\newblock \bibinfo{journal}{Trans. Assoc. Comput. Linguistics} \bibinfo{volume}{12} (\bibinfo{year}{2024}) \bibinfo{pages}{247--263}. \URLprefix \url{https://doi.org/10.1162/tacl\_a\_00646}. \DOIprefix\doi{10.1162/TACL\_A\_00646}.
\bibitem[{Lamurias et~al.(2020)Lamurias, Sousa, and Couto}]{BiQA}
\bibinfo{author}{A.~Lamurias}, \bibinfo{author}{D.~Sousa}, \bibinfo{author}{F.~M. Couto},
\newblock \bibinfo{title}{Generating biomedical question answering corpora from q\&a forums},
\newblock \bibinfo{journal}{IEEE Access} \bibinfo{volume}{8} (\bibinfo{year}{2020}) \bibinfo{pages}{161042--161051}. \DOIprefix\doi{10.1109/ACCESS.2020.3020868}.
\bibitem[{Zhu et~al.(2019)Zhu, Ahuja, Wei, and Reddy}]{HealthQA}
\bibinfo{author}{M.~Zhu}, \bibinfo{author}{A.~Ahuja}, \bibinfo{author}{W.~Wei}, \bibinfo{author}{C.~K. Reddy},
\newblock \bibinfo{title}{A hierarchical attention retrieval model for healthcare question answering},
\newblock in: \bibinfo{booktitle}{The World Wide Web Conference}, WWW '19, \bibinfo{publisher}{Association for Computing Machinery}, \bibinfo{address}{New York, NY, USA}, \bibinfo{year}{2019}, p. \bibinfo{pages}{2472–2482}. \URLprefix \url{https://doi.org/10.1145/3308558.3313699}. \DOIprefix\doi{10.1145/3308558.3313699}.
\bibitem[{Weinzierl and Harabagiu(2021)}]{Epic-QA}
\bibinfo{author}{M.~A. Weinzierl}, \bibinfo{author}{S.~M. Harabagiu},
\newblock \bibinfo{title}{The university of texas at dallas hltri's participation in {EPIC-QA:} searching for entailed questions revealing novel answer nuggets},
\newblock \bibinfo{journal}{CoRR} \bibinfo{volume}{abs/2112.13946} (\bibinfo{year}{2021}) \bibinfo{pages}{--}. \URLprefix \url{https://arxiv.org/abs/2112.13946}. \href{http://arxiv.org/abs/2112.13946}{{\tt arXiv:2112.13946}}.
\bibitem[{M{\"o}ller et~al.(2020)M{\"o}ller, Reina, Jayakumar, and Pietsch}]{moller2020covidqa}
\bibinfo{author}{T.~M{\"o}ller}, \bibinfo{author}{A.~Reina}, \bibinfo{author}{R.~Jayakumar}, \bibinfo{author}{M.~Pietsch},
\newblock \bibinfo{title}{{COVID}-{QA}: A question answering dataset for {COVID}-19},
\newblock in: \bibinfo{booktitle}{ACL 2020 Workshop on Natural Language Processing for COVID-19 (NLP-COVID)}, \bibinfo{publisher}{ACL}, \bibinfo{address}{Online}, \bibinfo{year}{2020}, pp.~\bibinfo{pages}{--}. \URLprefix \url{https://openreview.net/forum?id=JENSKEEzsoU}.
\bibitem[{{\v{S}}uster and Daelemans(2018)}]{suster-daelemans-2018-clicr}
\bibinfo{author}{S.~{\v{S}}uster}, \bibinfo{author}{W.~Daelemans},
\newblock \bibinfo{title}{{C}li{CR}: a dataset of clinical case reports for machine reading comprehension},
\newblock in: \bibinfo{editor}{M.~Walker}, \bibinfo{editor}{H.~Ji}, \bibinfo{editor}{A.~Stent} (Eds.), \bibinfo{booktitle}{Proceedings of the 2018 Conference of the North {A}merican Chapter of the Association for Computational Linguistics: Human Language Technologies, Volume 1 (Long Papers)}, \bibinfo{publisher}{Association for Computational Linguistics}, \bibinfo{address}{New Orleans, Louisiana}, \bibinfo{year}{2018}, pp. \bibinfo{pages}{1551--1563}. \URLprefix \url{https://aclanthology.org/N18-1140/}. \DOIprefix\doi{10.18653/v1/N18-1140}.
\bibitem[{Abacha et~al.(2017)Abacha, Agichtein, Pinter, and Demner{-}Fushman}]{LiveMedQA2017}
\bibinfo{author}{A.~B. Abacha}, \bibinfo{author}{E.~Agichtein}, \bibinfo{author}{Y.~Pinter}, \bibinfo{author}{D.~Demner{-}Fushman},
\newblock \bibinfo{title}{Overview of the medical question answering task at {TREC} 2017 liveqa},
\newblock in: \bibinfo{editor}{E.~M. Voorhees}, \bibinfo{editor}{A.~Ellis} (Eds.), \bibinfo{booktitle}{Proceedings of The Twenty-Sixth Text REtrieval Conference, {TREC} 2017, Gaithersburg, Maryland, USA, November 15-17, 2017}, volume \bibinfo{volume}{500-324} of \textit{\bibinfo{series}{{NIST} Special Publication}}, \bibinfo{publisher}{National Institute of Standards and Technology {(NIST)}}, \bibinfo{address}{Gaithersburg, Maryland, USA}, \bibinfo{year}{2017}, pp.~\bibinfo{pages}{--}. \URLprefix \url{https://trec.nist.gov/pubs/trec26/papers/Overview-QA.pdf}.
\bibitem[{Jin et~al.(2020)Jin, Pan, Oufattole, Weng, Fang, and Szolovits}]{medqa}
\bibinfo{author}{D.~Jin}, \bibinfo{author}{E.~Pan}, \bibinfo{author}{N.~Oufattole}, \bibinfo{author}{W.-H. Weng}, \bibinfo{author}{H.~Fang}, \bibinfo{author}{P.~Szolovits}, \bibinfo{title}{What disease does this patient have? a large-scale open domain question answering dataset from medical exams}, \bibinfo{year}{2020}. \URLprefix \url{https://arxiv.org/abs/2009.13081}. \href{http://arxiv.org/abs/2009.13081}{{\tt arXiv:2009.13081}}.
\bibitem[{Pampari et~al.(2018)Pampari, Raghavan, Liang, and Peng}]{pampari2018emrqa}
\bibinfo{author}{A.~Pampari}, \bibinfo{author}{P.~Raghavan}, \bibinfo{author}{J.~J. Liang}, \bibinfo{author}{J.~Peng},
\newblock \bibinfo{title}{emrqa: {A} large corpus for question answering on electronic medical records},
\newblock in: \bibinfo{editor}{E.~Riloff}, \bibinfo{editor}{D.~Chiang}, \bibinfo{editor}{J.~Hockenmaier}, \bibinfo{editor}{J.~Tsujii} (Eds.), \bibinfo{booktitle}{Proceedings of the 2018 Conference on Empirical Methods in Natural Language Processing, Brussels, Belgium, October 31 - November 4, 2018}, \bibinfo{publisher}{Association for Computational Linguistics}, \bibinfo{address}{Brussels, Belgium}, \bibinfo{year}{2018}, pp. \bibinfo{pages}{2357--2368}. \URLprefix \url{https://doi.org/10.18653/v1/d18-1258}. \DOIprefix\doi{10.18653/V1/D18-1258}.
\bibitem[{He et~al.(2019)He, Fu, and Tu}]{webmedqa}
\bibinfo{author}{J.~He}, \bibinfo{author}{M.~Fu}, \bibinfo{author}{M.~Tu},
\newblock \bibinfo{title}{Applying deep matching networks to chinese medical question answering: a study and a dataset},
\newblock \bibinfo{journal}{{BMC} Medical Informatics Decis. Mak.} \bibinfo{volume}{19-S} (\bibinfo{year}{2019}) \bibinfo{pages}{91--100}. \URLprefix \url{https://doi.org/10.1186/s12911-019-0761-8}. \DOIprefix\doi{10.1186/S12911-019-0761-8}.
\bibitem[{Nentidis et~al.(2020)Nentidis, Krithara, Bougiatiotis, Krallinger, Penagos, Villegas, and Paliouras}]{bioasq}
\bibinfo{author}{A.~Nentidis}, \bibinfo{author}{A.~Krithara}, \bibinfo{author}{K.~Bougiatiotis}, \bibinfo{author}{M.~Krallinger}, \bibinfo{author}{C.~R. Penagos}, \bibinfo{author}{M.~Villegas}, \bibinfo{author}{G.~Paliouras},
\newblock \bibinfo{title}{Overview of bioasq 2020: The eighth bioasq challenge on large-scale biomedical semantic indexing and question answering},
\newblock in: \bibinfo{editor}{A.~Arampatzis}, \bibinfo{editor}{E.~Kanoulas}, \bibinfo{editor}{T.~Tsikrika}, \bibinfo{editor}{S.~Vrochidis}, \bibinfo{editor}{H.~Joho}, \bibinfo{editor}{C.~Lioma}, \bibinfo{editor}{C.~Eickhoff}, \bibinfo{editor}{A.~N{\'{e}}v{\'{e}}ol}, \bibinfo{editor}{L.~Cappellato}, \bibinfo{editor}{N.~Ferro} (Eds.), \bibinfo{booktitle}{Experimental {IR} Meets Multilinguality, Multimodality, and Interaction - 11th International Conference of the {CLEF} Association, {CLEF} 2020, Thessaloniki, Greece, September 22-25, 2020, Proceedings}, volume \bibinfo{volume}{12260} of \textit{\bibinfo{series}{Lecture Notes in Computer Science}}, \bibinfo{publisher}{Springer}, \bibinfo{address}{Thessaloniki, Greece}, \bibinfo{year}{2020}, pp. \bibinfo{pages}{194--214}. \URLprefix \url{https://doi.org/10.1007/978-3-030-58219-7\_16}. \DOIprefix\doi{10.1007/978-3-030-58219-7\_16}.
\bibitem[{Vilares and G{\'o}mez-Rodr{\'i}guez(2019)}]{vilares-gomez-rodriguez-2019-head}
\bibinfo{author}{D.~Vilares}, \bibinfo{author}{C.~G{\'o}mez-Rodr{\'i}guez},
\newblock \bibinfo{title}{{HEAD}-{QA}: A healthcare dataset for complex reasoning},
\newblock in: \bibinfo{editor}{A.~Korhonen}, \bibinfo{editor}{D.~Traum}, \bibinfo{editor}{L.~M{\`a}rquez} (Eds.), \bibinfo{booktitle}{Proceedings of the 57th Annual Meeting of the Association for Computational Linguistics}, \bibinfo{publisher}{Association for Computational Linguistics}, \bibinfo{address}{Florence, Italy}, \bibinfo{year}{2019}, pp. \bibinfo{pages}{960--966}. \URLprefix \url{https://aclanthology.org/P19-1092/}. \DOIprefix\doi{10.18653/v1/P19-1092}.
\bibitem[{Pal et~al.(2022)Pal, Umapathi, and Sankarasubbu}]{MedMCQA}
\bibinfo{author}{A.~Pal}, \bibinfo{author}{L.~K. Umapathi}, \bibinfo{author}{M.~Sankarasubbu},
\newblock \bibinfo{title}{Medmcqa: {A} large-scale multi-subject multi-choice dataset for medical domain question answering},
\newblock in: \bibinfo{editor}{G.~Flores}, \bibinfo{editor}{G.~H. Chen}, \bibinfo{editor}{T.~J. Pollard}, \bibinfo{editor}{J.~C. Ho}, \bibinfo{editor}{T.~Naumann} (Eds.), \bibinfo{booktitle}{Conference on Health, Inference, and Learning, {CHIL} 2022, 7-8 April 2022, Virtual Event}, volume \bibinfo{volume}{174} of \textit{\bibinfo{series}{Proceedings of Machine Learning Research}}, \bibinfo{publisher}{{PMLR}}, \bibinfo{address}{Online}, \bibinfo{year}{2022}, pp. \bibinfo{pages}{248--260}. \URLprefix \url{https://proceedings.mlr.press/v174/pal22a.html}.
\bibitem[{Tian et~al.(2019)Tian, Ma, Xia, and Song}]{tian-etal-2019-chimed}
\bibinfo{author}{Y.~Tian}, \bibinfo{author}{W.~Ma}, \bibinfo{author}{F.~Xia}, \bibinfo{author}{Y.~Song},
\newblock \bibinfo{title}{{C}hi{M}ed: A {C}hinese medical corpus for question answering},
\newblock in: \bibinfo{editor}{D.~Demner-Fushman}, \bibinfo{editor}{K.~B. Cohen}, \bibinfo{editor}{S.~Ananiadou}, \bibinfo{editor}{J.~Tsujii} (Eds.), \bibinfo{booktitle}{Proceedings of the 18th BioNLP Workshop and Shared Task}, \bibinfo{publisher}{Association for Computational Linguistics}, \bibinfo{address}{Florence, Italy}, \bibinfo{year}{2019}, pp. \bibinfo{pages}{250--260}. \URLprefix \url{https://aclanthology.org/W19-5027/}. \DOIprefix\doi{10.18653/v1/W19-5027}.
\bibitem[{Pe{\~{n}}as et~al.(2013)Pe{\~{n}}as, Hovy, Forner, Rodrigo, Sutcliffe, and Morante}]{Peas2013QA4MRE2O}
\bibinfo{author}{A.~Pe{\~{n}}as}, \bibinfo{author}{E.~H. Hovy}, \bibinfo{author}{P.~Forner}, \bibinfo{author}{{\'{A}}.~Rodrigo}, \bibinfo{author}{R.~F.~E. Sutcliffe}, \bibinfo{author}{R.~Morante},
\newblock \bibinfo{title}{{QA4MRE} 2011-2013: Overview of question answering for machine reading evaluation},
\newblock in: \bibinfo{editor}{P.~Forner}, \bibinfo{editor}{H.~M{\"{u}}ller}, \bibinfo{editor}{R.~Paredes}, \bibinfo{editor}{P.~Rosso}, \bibinfo{editor}{B.~Stein} (Eds.), \bibinfo{booktitle}{Information Access Evaluation. Multilinguality, Multimodality, and Visualization - 4th International Conference of the {CLEF} Initiative, {CLEF} 2013, Valencia, Spain, September 23-26, 2013. Proceedings}, volume \bibinfo{volume}{8138} of \textit{\bibinfo{series}{Lecture Notes in Computer Science}}, \bibinfo{publisher}{Springer}, \bibinfo{address}{Valencia, Spain}, \bibinfo{year}{2013}, pp. \bibinfo{pages}{303--320}. \URLprefix \url{https://doi.org/10.1007/978-3-642-40802-1\_29}. \DOIprefix\doi{10.1007/978-3-642-40802-1\_29}.
\bibitem[{Pappas et~al.(2018)Pappas, Androutsopoulos, and Papageorgiou}]{pappas-etal-2018-bioread}
\bibinfo{author}{D.~Pappas}, \bibinfo{author}{I.~Androutsopoulos}, \bibinfo{author}{H.~Papageorgiou},
\newblock \bibinfo{title}{Bioread: {A} new dataset for biomedical reading comprehension},
\newblock in: \bibinfo{editor}{N.~Calzolari}, \bibinfo{editor}{K.~Choukri}, \bibinfo{editor}{C.~Cieri}, \bibinfo{editor}{T.~Declerck}, \bibinfo{editor}{S.~Goggi}, \bibinfo{editor}{K.~Hasida}, \bibinfo{editor}{H.~Isahara}, \bibinfo{editor}{B.~Maegaard}, \bibinfo{editor}{J.~Mariani}, \bibinfo{editor}{H.~Mazo}, \bibinfo{editor}{A.~Moreno}, \bibinfo{editor}{J.~Odijk}, \bibinfo{editor}{S.~Piperidis}, \bibinfo{editor}{T.~Tokunaga} (Eds.), \bibinfo{booktitle}{Proceedings of the Eleventh International Conference on Language Resources and Evaluation, {LREC} 2018, Miyazaki, Japan, May 7-12, 2018}, \bibinfo{publisher}{European Language Resources Association {(ELRA)}}, \bibinfo{address}{Miyazaki, Japan}, \bibinfo{year}{2018}, pp.~\bibinfo{pages}{--}. \URLprefix \url{http://www.lrec-conf.org/proceedings/lrec2018/summaries/795.html}.
\bibitem[{Pappas et~al.(2020)Pappas, Stavropoulos, Androutsopoulos, and McDonald}]{pappas-etal-2020-biomrc}
\bibinfo{author}{D.~Pappas}, \bibinfo{author}{P.~Stavropoulos}, \bibinfo{author}{I.~Androutsopoulos}, \bibinfo{author}{R.~McDonald},
\newblock \bibinfo{title}{{B}io{MRC}: A dataset for biomedical machine reading comprehension},
\newblock in: \bibinfo{editor}{D.~Demner-Fushman}, \bibinfo{editor}{K.~B. Cohen}, \bibinfo{editor}{S.~Ananiadou}, \bibinfo{editor}{J.~Tsujii} (Eds.), \bibinfo{booktitle}{Proceedings of the 19th SIGBioMed Workshop on Biomedical Language Processing}, \bibinfo{publisher}{Association for Computational Linguistics}, \bibinfo{address}{Online}, \bibinfo{year}{2020}, pp. \bibinfo{pages}{140--149}. \URLprefix \url{https://aclanthology.org/2020.bionlp-1.15/}. \DOIprefix\doi{10.18653/v1/2020.bionlp-1.15}.
\bibitem[{Christophe et~al.(2024)Christophe, Kanithi, Raha, Khan, and Pimentel}]{llmMed42}
\bibinfo{author}{C.~Christophe}, \bibinfo{author}{P.~K. Kanithi}, \bibinfo{author}{T.~Raha}, \bibinfo{author}{S.~Khan}, \bibinfo{author}{M.~A. Pimentel},
\newblock \bibinfo{title}{Med42-v2: {A} suite of clinical llms},
\newblock \bibinfo{journal}{CoRR} \bibinfo{volume}{abs/2408.06142} (\bibinfo{year}{2024}) \bibinfo{pages}{--}. \URLprefix \url{https://doi.org/10.48550/arXiv.2408.06142}. \DOIprefix\doi{10.48550/ARXIV.2408.06142}. \href{http://arxiv.org/abs/2408.06142}{{\tt arXiv:2408.06142}}.
\bibitem[{DeepMount00(2024)}]{ItalianNERXXL}
\bibinfo{author}{DeepMount00}, \bibinfo{title}{Italian\_ner\_xxl}, \bibinfo{howpublished}{https://huggingface.co/DeepMount00/Italian\_NER\_XXL}, \bibinfo{year}{2024}.
\bibitem[{Grootendorst(2022)}]{Berttopic}
\bibinfo{author}{M.~Grootendorst},
\newblock \bibinfo{title}{Bertopic: Neural topic modeling with a class-based {TF-IDF} procedure},
\newblock \bibinfo{journal}{CoRR} \bibinfo{volume}{abs/2203.05794} (\bibinfo{year}{2022}) \bibinfo{pages}{--}. \URLprefix \url{https://doi.org/10.48550/arXiv.2203.05794}. \DOIprefix\doi{10.48550/ARXIV.2203.05794}. \href{http://arxiv.org/abs/2203.05794}{{\tt arXiv:2203.05794}}.
\bibitem[{Reimers and Gurevych(2019)}]{reimers-2019-sentence-bert}
\bibinfo{author}{N.~Reimers}, \bibinfo{author}{I.~Gurevych},
\newblock \bibinfo{title}{Sentence-bert: Sentence embeddings using siamese bert-networks},
\newblock in: \bibinfo{editor}{K.~Inui}, \bibinfo{editor}{J.~Jiang}, \bibinfo{editor}{V.~Ng}, \bibinfo{editor}{X.~Wan} (Eds.), \bibinfo{booktitle}{Proceedings of the 2019 Conference on Empirical Methods in Natural Language Processing and the 9th International Joint Conference on Natural Language Processing, {EMNLP-IJCNLP} 2019, Hong Kong, China, November 3-7, 2019}, \bibinfo{publisher}{Association for Computational Linguistics}, \bibinfo{address}{Hong Kong, China}, \bibinfo{year}{2019}, pp. \bibinfo{pages}{3980--3990}. \URLprefix \url{https://doi.org/10.18653/v1/D19-1410}. \DOIprefix\doi{10.18653/V1/D19-1410}.
\bibitem[{McInnes and Healy(2018)}]{UMAP}
\bibinfo{author}{L.~McInnes}, \bibinfo{author}{J.~Healy},
\newblock \bibinfo{title}{{UMAP:} uniform manifold approximation and projection for dimension reduction},
\newblock \bibinfo{journal}{CoRR} \bibinfo{volume}{abs/1802.03426} (\bibinfo{year}{2018}) \bibinfo{pages}{--}. \URLprefix \url{http://arxiv.org/abs/1802.03426}. \href{http://arxiv.org/abs/1802.03426}{{\tt arXiv:1802.03426}}.
\bibitem[{Rahman et~al.(2016)Rahman, Liu, Suhaim, Thirumuruganathan, Zhang, and Das}]{HDBSCAN}
\bibinfo{author}{M.~F. Rahman}, \bibinfo{author}{W.~Liu}, \bibinfo{author}{S.~B. Suhaim}, \bibinfo{author}{S.~Thirumuruganathan}, \bibinfo{author}{N.~Zhang}, \bibinfo{author}{G.~Das},
\newblock \bibinfo{title}{{HDBSCAN:} density based clustering over location based services},
\newblock \bibinfo{journal}{CoRR} \bibinfo{volume}{abs/1602.03730} (\bibinfo{year}{2016}) \bibinfo{pages}{--}. \URLprefix \url{http://arxiv.org/abs/1602.03730}. \href{http://arxiv.org/abs/1602.03730}{{\tt arXiv:1602.03730}}.
\bibitem[{Liu et~al.(2023)Liu, Zhou, Hua, Chong, Tian, Liu, Wang, You, Guo, Zhu, and Li}]{liu2023nips}
\bibinfo{author}{J.~Liu}, \bibinfo{author}{P.~Zhou}, \bibinfo{author}{Y.~Hua}, \bibinfo{author}{D.~Chong}, \bibinfo{author}{Z.~Tian}, \bibinfo{author}{A.~Liu}, \bibinfo{author}{H.~Wang}, \bibinfo{author}{C.~You}, \bibinfo{author}{Z.~Guo}, \bibinfo{author}{L.~Zhu}, \bibinfo{author}{M.~L. Li},
\newblock \bibinfo{title}{Benchmarking large language models on cmexam - {A} comprehensive chinese medical exam dataset},
\newblock in: \bibinfo{editor}{A.~Oh}, \bibinfo{editor}{T.~Naumann}, \bibinfo{editor}{A.~Globerson}, \bibinfo{editor}{K.~Saenko}, \bibinfo{editor}{M.~Hardt}, \bibinfo{editor}{S.~Levine} (Eds.), \bibinfo{booktitle}{Advances in Neural Information Processing Systems 36: Annual Conference on Neural Information Processing Systems 2023, NeurIPS 2023, New Orleans, LA, USA, December 10 - 16, 2023}, \bibinfo{publisher}{NeurIPS}, \bibinfo{address}{New Orleans, LA, USA}, \bibinfo{year}{2023}, pp.~\bibinfo{pages}{--}. \URLprefix \url{http://papers.nips.cc/paper\_files/paper/2023/hash/a48ad12d588c597f4725a8b84af647b5-Abstract-Datasets\_and\_Benchmarks.html}.
\bibitem[{Jin et~al.(2024)Jin, Chandra, Verma, Hu, Choudhury, and Kumar}]{jin2024www}
\bibinfo{author}{Y.~Jin}, \bibinfo{author}{M.~Chandra}, \bibinfo{author}{G.~Verma}, \bibinfo{author}{Y.~Hu}, \bibinfo{author}{M.~D. Choudhury}, \bibinfo{author}{S.~Kumar},
\newblock \bibinfo{title}{Better to ask in english: Cross-lingual evaluation of large language models for healthcare queries},
\newblock in: \bibinfo{editor}{T.~Chua}, \bibinfo{editor}{C.~Ngo}, \bibinfo{editor}{R.~Kumar}, \bibinfo{editor}{H.~W. Lauw}, \bibinfo{editor}{R.~K. Lee} (Eds.), \bibinfo{booktitle}{Proceedings of the {ACM} on Web Conference 2024, {WWW} 2024, Singapore, May 13-17, 2024}, \bibinfo{publisher}{{ACM}}, \bibinfo{address}{Singapore}, \bibinfo{year}{2024}, pp. \bibinfo{pages}{2627--2638}. \URLprefix \url{https://doi.org/10.1145/3589334.3645643}. \DOIprefix\doi{10.1145/3589334.3645643}.
\bibitem[{Zhang et~al.(2020)Zhang, Kishore, Wu, Weinberger, and Artzi}]{zhang2020iclr}
\bibinfo{author}{T.~Zhang}, \bibinfo{author}{V.~Kishore}, \bibinfo{author}{F.~Wu}, \bibinfo{author}{K.~Q. Weinberger}, \bibinfo{author}{Y.~Artzi},
\newblock \bibinfo{title}{Bertscore: Evaluating text generation with {BERT}},
\newblock in: \bibinfo{booktitle}{8th International Conference on Learning Representations, {ICLR} 2020, Addis Ababa, Ethiopia, April 26-30, 2020}, \bibinfo{publisher}{OpenReview.net}, \bibinfo{address}{Addis Ababa, Ethiopia}, \bibinfo{year}{2020}, pp.~\bibinfo{pages}{--}. \URLprefix \url{https://openreview.net/forum?id=SkeHuCVFDr}.
\bibitem[{Zhang et~al.(2017)Zhang, Zhang, Wang, Cheng, Li, and Ding}]{zhang2017cMedQA}
\bibinfo{author}{S.~Zhang}, \bibinfo{author}{X.~Zhang}, \bibinfo{author}{H.~Wang}, \bibinfo{author}{J.~Cheng}, \bibinfo{author}{P.~Li}, \bibinfo{author}{Z.~Ding},
\newblock \bibinfo{title}{Chinese medical question answer matching using end-to-end character-level multi-scale cnns},
\newblock \bibinfo{journal}{Applied Sciences} \bibinfo{volume}{7} (\bibinfo{year}{2017}) \bibinfo{pages}{767}.
\bibitem[{Yagnik et~al.(2024)Yagnik, Jhaveri, Sharma, Pila, Ben, and Shang}]{DBLP:journals/corr/abs-2401-11389}
\bibinfo{author}{N.~Yagnik}, \bibinfo{author}{J.~Jhaveri}, \bibinfo{author}{V.~Sharma}, \bibinfo{author}{G.~Pila}, \bibinfo{author}{A.~Ben}, \bibinfo{author}{J.~Shang},
\newblock \bibinfo{title}{Medlm: Exploring language models for medical question answering systems},
\newblock \bibinfo{journal}{CoRR} \bibinfo{volume}{abs/2401.11389} (\bibinfo{year}{2024}) \bibinfo{pages}{--}. \URLprefix \url{https://doi.org/10.48550/arXiv.2401.11389}. \DOIprefix\doi{10.48550/ARXIV.2401.11389}. \href{http://arxiv.org/abs/2401.11389}{{\tt arXiv:2401.11389}}.
\bibitem[{Douze et~al.(2024)Douze, Guzhva, Deng, Johnson, Szilvasy, Mazaré, Lomeli, Hosseini, and Jégou}]{douze2024faiss}
\bibinfo{author}{M.~Douze}, \bibinfo{author}{A.~Guzhva}, \bibinfo{author}{C.~Deng}, \bibinfo{author}{J.~Johnson}, \bibinfo{author}{G.~Szilvasy}, \bibinfo{author}{P.-E. Mazaré}, \bibinfo{author}{M.~Lomeli}, \bibinfo{author}{L.~Hosseini}, \bibinfo{author}{H.~Jégou},
\newblock \bibinfo{title}{The faiss library},
\newblock \bibinfo{journal}{arXiv preprint arXiv:2401.08281}  (\bibinfo{year}{2024}). \href{http://arxiv.org/abs/2401.08281}{{\tt arXiv:2401.08281}}.
\bibitem[{Tran et~al.(2024)Tran, Yang, Yao, and Yu}]{tran2024jamia}
\bibinfo{author}{H.~Tran}, \bibinfo{author}{Z.~Yang}, \bibinfo{author}{Z.~Yao}, \bibinfo{author}{H.~Yu},
\newblock \bibinfo{title}{Bioinstruct: instruction tuning of large language models for biomedical natural language processing},
\newblock \bibinfo{journal}{J. Am. Medical Informatics Assoc.} \bibinfo{volume}{31} (\bibinfo{year}{2024}) \bibinfo{pages}{1821--1832}. \URLprefix \url{https://doi.org/10.1093/jamia/ocae122}. \DOIprefix\doi{10.1093/JAMIA/OCAE122}.
\bibitem[{Dorfner et~al.(2024)Dorfner, Dada, Busch, Makowski, Han, Truhn, Kleesiek, Sushil, Lammert, Adams, and Bressem}]{DBLP:journals/corr/abs-2408-13833}
\bibinfo{author}{F.~J. Dorfner}, \bibinfo{author}{A.~Dada}, \bibinfo{author}{F.~Busch}, \bibinfo{author}{M.~R. Makowski}, \bibinfo{author}{T.~Han}, \bibinfo{author}{D.~Truhn}, \bibinfo{author}{J.~Kleesiek}, \bibinfo{author}{M.~Sushil}, \bibinfo{author}{J.~Lammert}, \bibinfo{author}{L.~C. Adams}, \bibinfo{author}{K.~K. Bressem},
\newblock \bibinfo{title}{Biomedical large languages models seem not to be superior to generalist models on unseen medical data},
\newblock \bibinfo{journal}{CoRR} \bibinfo{volume}{abs/2408.13833} (\bibinfo{year}{2024}) \bibinfo{pages}{--}. \URLprefix \url{https://doi.org/10.48550/arXiv.2408.13833}. \DOIprefix\doi{10.48550/ARXIV.2408.13833}. \href{http://arxiv.org/abs/2408.13833}{{\tt arXiv:2408.13833}}.
\bibitem[{Van~Nguyen et~al.(2024)Van~Nguyen, Shen, Aponte, Xia, Basu, Hu, Chen, Parmar, Kunapuli, Barrow et~al.}]{van2024survey}
\bibinfo{author}{C.~Van~Nguyen}, \bibinfo{author}{X.~Shen}, \bibinfo{author}{R.~Aponte}, \bibinfo{author}{Y.~Xia}, \bibinfo{author}{S.~Basu}, \bibinfo{author}{Z.~Hu}, \bibinfo{author}{J.~Chen}, \bibinfo{author}{M.~Parmar}, \bibinfo{author}{S.~Kunapuli}, \bibinfo{author}{J.~Barrow}, et~al.,
\newblock \bibinfo{title}{A survey of small language models},
\newblock \bibinfo{journal}{arXiv preprint arXiv:2410.20011}  (\bibinfo{year}{2024}).
\bibitem[{Bengio et~al.(2009)Bengio, Louradour, Collobert, and Weston}]{curriculum10.1145/1553374.1553380}
\bibinfo{author}{Y.~Bengio}, \bibinfo{author}{J.~Louradour}, \bibinfo{author}{R.~Collobert}, \bibinfo{author}{J.~Weston},
\newblock \bibinfo{title}{Curriculum learning},
\newblock in: \bibinfo{booktitle}{Proceedings of the 26th Annual International Conference on Machine Learning}, ICML '09, \bibinfo{publisher}{Association for Computing Machinery}, \bibinfo{address}{New York, NY, USA}, \bibinfo{year}{2009}, p. \bibinfo{pages}{41–48}. \URLprefix \url{https://doi.org/10.1145/1553374.1553380}. \DOIprefix\doi{10.1145/1553374.1553380}.
\bibitem[{Daniel~Han and team(2023)}]{unsloth}
\bibinfo{author}{M.~H. Daniel~Han}, \bibinfo{author}{U.~team}, \bibinfo{title}{Unsloth}, \bibinfo{year}{2023}. \URLprefix \url{http://github.com/unslothai/unsloth}.

\end{thebibliography}


\end{document}